\documentclass[journal,twoside,web]{ieeecolor2}
\usepackage{generic}
\usepackage{cite}
\usepackage{amsmath,amssymb,amsfonts}
\usepackage{graphicx}
\usepackage{amsmath,amsfonts}
\usepackage{graphicx}
\usepackage{subcaption}
\usepackage{booktabs}
\usepackage{multirow}
\usepackage{tabularx}
\usepackage{colortbl}
\usepackage{textcomp}
\usepackage{algorithm}
\usepackage{algpseudocode}
\usepackage{siunitx}
\usepackage{graphicx}
\usepackage{subcaption}
\usepackage{textcomp}

\usepackage[utf8]{inputenc}

\usepackage{booktabs} % For better table formatting
\usepackage{longtable} % For tables that span multiple pages, if needed

\def\BibTeX{{\rm B\kern-.05em{\sc i\kern-.025em b}\kern-.08em
    T\kern-.1667em\lower.7ex\hbox{E}\kern-.125emX}}
\begin{document}
\title{Robust Support Vector Machines for Imbalanced and Noisy Data via Benders Decomposition}
\author{Seyed Mojtaba Mohasel, Hamidreza Koosha
\thanks{Seyed Mojtaba Mohasel is with the Department of Mechanical and Industrial Engineering, Montana State University, Montana, USA (e-mail: Seyedmojtabamohasel@montana.edu).}
\thanks{Hamidreza Koosha is with the Department of Industrial Engineering, Ferdowsi University of Mashhad, Mashhad, Iran (e-mail: koosha@um.ac.ir).}
}

\maketitle

\begin{abstract} 

This study introduces a novel formulation to enhance Support Vector Machines (SVMs) in handling class imbalance and noise. Unlike the conventional Soft Margin SVM, which penalizes the magnitude of constraint violations, the proposed model quantifies the number of violations and aims to minimize their frequency. To achieve this, a binary variable is incorporated into the objective function of the primal SVM formulation, replacing the traditional slack variable. Furthermore, each misclassified sample is assigned a priority and an associated constraint. The resulting formulation is a mixed-integer programming model, efficiently solved using Benders decomposition. The proposed model's performance was benchmarked against existing models, including Soft Margin SVM, weighted SVM, and NuSVC. Two primary hypotheses were examined: 1) The proposed model improves the F1-score for the minority class in imbalanced classification tasks. 2) The proposed model enhances classification accuracy in noisy datasets.
These hypotheses were evaluated using a Wilcoxon test across multiple publicly available datasets from the OpenML repository. The results supported both hypotheses (\( p < 0.05 \)). In addition, the proposed model exhibited several interesting properties, such as improved robustness to noise, a decision boundary shift favoring the minority class, a reduced number of support vectors, and decreased prediction time. The open-source Python implementation of the proposed SVM model is available.

\end{abstract}


\section{Introduction} 
Support Vector Machine (SVM) is a supervised machine learning (ML) algorithm with a model-based approach \cite{geron2022hands}. It is widely applied to classification tasks across various domains, including biomechanics \cite{wu2017detection}, bioinformatics \cite{byvatov2003support}, healthcare \cite{venkatesan2018ecg}, finance \cite{tay2001application}, marketing \cite{cui2005prediction}, image classification \cite{chen2014applying}, and audio classification \cite{grama2017optimization}. SVM's strong theoretical foundation, ability to yield optimal solutions, and remarkable generalization performance have made it highly valuable in data mining, pattern recognition, and machine learning research \cite{cervantes2020comprehensive}.

Hard margin SVM is formulated under the assumption that the classes are perfectly separable. The margin, which SVM enforces, is the maximized distance between the classes. However, in many real-world problems, the assumption of perfectly separable classes does not hold. For non-separable cases, Soft Margin SVM was introduced, which allows some violations of the margin constraints. Soft Margin SVM utilizes slack variables to quantify these violations, permitting some data points to be within the margin or even misclassified \cite{liu2025support}.

Soft Margin SVM classifies samples by constructing a boundary or hyperplane. It assumes that the best decision boundary between classes is the one that maximizes the margin, which indirectly reduces the risk of misclassification. Soft Margin SVM transforms the primal optimization problem into a dual problem and solves it as a quadratic programming problem. The solution to this problem identifies the support vectors, which define the decision boundary. Despite its strong mathematical foundation and generalization capacity for class-balanced datasets, Soft Margin SVM can perform poorly in cases of: 1) class imbalance \cite{rezvani2023broad,cervantes2020comprehensive,tanveer2025enhancing} or 2) noisy datasets \cite{xia2015relative, akhtar2024gl}.

Class imbalance occurs when the number of samples in each class is uneven. The class with more samples is called the majority class, while the class with fewer samples is called the minority class. ML models assume an equal class distribution; therefore, they typically perform poorly on the minority class. Class imbalance issue is commonly found in numerous applications, including detecting fraud or intrusions, managing risks, classifying text, diagnosing or monitoring medical conditions \cite{chawla2004special}, and identifying oil spills from radar images of the sea surface \cite{sun2009classification}. 

Class imbalance is commonly measured using the imbalance ratio (IR), calculated as the number of majority samples divided by the number of minority samples \cite{santos2022joint}. The IR can range from 1.25 \cite{gosain2017handling} to 10000 \cite{sun2009classification,ribeiro2020imbalanced} or more. Factors affecting the problem severity include sample size, class separability, and within-class concepts  \cite{sun2009classification}.

Methods for handling class imbalance can be categorized into data level, algorithmic level, and hybrid approaches \cite{krawczyk2016learning}. 
\begin{enumerate}
    \item Data level solutions involve using sampling strategies to balance the dataset and train the classifier with the balanced data. Commonly used approaches are oversampling \cite{islam2022knnor}, \cite{gosain2017handling}, \cite{saez2016analyzing} the minority class by generating synthetic samples \cite{chawla2002smote} or undersampling the majority class \cite{devi2020review}. However, sampling strategies are less effective for SVMs since SVMs calculate boundaries based only on support vectors, and class sizes may not significantly influence the class boundary \cite{sun2009classification}.

    \item Algorithmic-level solutions modify the ML algorithm by assigning different costs \cite{haixiang2017learning} to each class to address class imbalance. Developed algorithms include Weighted SVM \cite{chang2011libsvm}; however, determining the appropriate weights remains a challenge. In addition, it is not clear whether assigning weights can effectively achieve higher performance for the minority class. Therefore, a method is needed to effectively determine the decision boundary in favor of the minority class, particularly when the minority class is the primary concern for the end user (\textbf{Knowledge gap 1}).
    
    \item Hybrid approaches combine sampling strategies and algorithm modifications to handle class imbalance \cite{haixiang2017learning}. These approaches offer greater flexibility in handling class imbalance and can adapt to varying levels of imbalance ratios. However, balancing the trade-off between oversampling and undersampling can be challenging. Furthermore, hybrid approaches are inherently more complex to implement compared to methods focused solely on either data-level or algorithmic-level solutions. 
    \end{enumerate}

Noise is defined as erroneous or random data points. Noise can appear in multiple forms, including outliers, feature noise, and label noise
 \cite{kumar2024depth}. An outlier is a data point that significantly deviates from the norm of a dataset. Feature noise affects
the feature values of each sample.  Label noise refers to samples that are wrongly labeled as another class.

The Soft Margin SVM has serious shortcomings with the noisy datasets. The presence of noisy data (outliers) plays a significant role in determining the decision hyperplane, as these points tend to have the largest margin loss \cite{xu2006robust}. A robust margin loss, which does not increase the penalty beyond a certain point, was introduced as a remedy to offset the effect of noise in Soft Margin  SVM \cite{bartlett2002rademacher} \cite{krause2004leveraging} \cite{mason1999functional}. A drawback of these methods is that they compromise the convexity of the training objective, making global optimization unattainable \cite{xu2006robust}. Therefore, a method is needed to preserve the convexity of the objective function and enhance the robustness of SVM against noise (\textbf{Knowledge gap 2}).

The limitations of the existing models motivated this study to develop a novel SVM formulation. The Soft Margin SVM incorporates a slack variable to penalize margin violations based on their severity. However, in class-imbalanced datasets, penalizing misclassifications of minority and majority class samples equally can lead to suboptimal decision boundaries. Furthermore, in noisy datasets, outliers can distort optimization and lead to the selection of inappropriate support vectors.

To overcome these limitations and address the identified knowledge gaps, we hypothesized that an SVM model designed to minimize misclassification while maximizing the margin would demonstrate superior performance in handling class imbalance and noise compared to the conventional Soft Margin SVM. Based on this premise, we seek to answer the following research questions (RQ):

\textbf{RQ1}: Does moving the decision boundary in favor of the minority class improve SVM performance in class imbalance scenarios?

\textbf{RQ2}: Does gradually creating the decision boundary by updating support vectors improve performance when there is an overlap between classes?

Addressing these research questions leads to the development of a novel SVM formulation with key contributions. The proposed model offers:

\begin{itemize}
    \item Enhanced handling of class imbalance, particularly when the minority class is the primary focus for end users.
    \item Increased robustness against noise, ensuring reliable classification when all classes hold equal significance.
\end{itemize}

The remainder of this paper is organized as follows. 
Section II introduces the formulation of the proposed SVM model, which incorporates a mixed-integer programming structure. A mathematical programming framework with decomposition techniques is then employed to solve the model. To evaluate its effectiveness, the proposed model is compared against benchmark methods across multiple public datasets, focusing on scenarios with class imbalance and noise. 
Section {III} presents the results along with a comparative analysis of different methods. 
Section {IV} discusses the strengths and limitations of the model, highlighting its distinctions from benchmark methods. 
Finally, Section {V} concludes the paper.

\section{Proposed Method}
This section provides a brief overview of Soft Margin SVM and its variants. The proposed model is then introduced, with a focus on its distinctions from existing models.

\subsection{Soft Margin SVM} Soft Margin SVM aims to minimize the weight vector, which corresponds to maximizing the margin between the two classes. The margin is the distance between the hyperplane and the nearest data points from each class, known as support vectors. Additionally, it penalizes violations of the margin based on the extent of the violation.

\begin{equation}
    \min_{\mathbf{w}, b, \boldsymbol{\xi}} \quad \frac{1}{2} \mathbf{w}^T \mathbf{w} + C \sum_{i=1}^{n} \xi_i
    \label{eq:svm_objective}
\end{equation}

\textbf{Subject to:}

\begin{equation}
     \quad y_i (\mathbf{w}^T \mathbf{x}_i + b) \geq 1 - \xi_i,       \forall i
     \label{eq:svm_constraint}
\end{equation}

\begin{equation}
    \xi_i \geq 0
    \label{eq:svm_constraint2}
\end{equation}

In Equation \eqref{eq:svm_objective}, the weight vector \( \mathbf{w} \) characterizes the model’s orientation, determining the direction of the decision boundary. The slack variable \( \xi_i \) is introduced to handle Soft Margin classification by allowing some relaxation of the margin constraint; it quantifies the extent to which a data point violates the margin constraints. The regularization parameter \( C \) controls the trade-off between maximizing the margin and minimizing classification errors. 

In Equation \eqref{eq:svm_constraint}, the vector \( \mathbf{x}_i \) represents the \( i \)-th data point, and \( y_i \) is its corresponding class label, which takes values in \( \{-1,1\} \). The bias term \( b \) allows for shifting the hyperplane without altering its orientation. Constraint (3), ensures that the slack variables \( \xi_i \) are non-negative.

\subsection{Weighted SVM}
Weighted SVM is a variant of the Soft Margin SVM designed for imbalanced datasets.

\begin{equation}
    \min_{\mathbf{w}, b, \boldsymbol{\xi}} \quad \frac{1}{2} \mathbf{w}^T \mathbf{w} + C \sum_{i=1}^{n} w_{y_i} \xi_i
    \label{eq:svm_imbalance}
\end{equation}

In Equation \eqref{eq:svm_imbalance}, \( w_{y_i} \) is a class-specific weight, where a higher weight is assigned to the minority class to address class imbalance. The constraint remains the same as in the Soft Margin SVM formulation.

\subsection{NuSVC}
NuSVC is another variant of Soft Margin SVM that uses the hyperparameter \(\nu\), which directly bounds the fraction of margin errors (misclassifications) and the fraction of support vectors \cite{chang2011libsvm}.

\begin{equation}
\min_{\mathbf{w}, b, \rho, \boldsymbol{\xi}} \quad \frac{1}{2} \mathbf{w}^T \mathbf{w} - \nu \rho + \frac{1}{n} \sum_{i=1}^{n} \xi_i
\label{eq:nu_svm}
\end{equation}

\textbf{Subject to:}

\begin{equation}
y_i (\mathbf{w}^T \mathbf{x}_i+ b) \geq \rho - \xi_i, \quad \forall i
\label{eq:nu_svm_constraint1}
\end{equation}

\begin{equation}
\xi_i \geq 0, \quad \forall i
\label{eq:nu_svm_constraint2}
\end{equation}

\begin{equation}
\rho \geq 0.
\label{eq:nu_svm_constraint3}
\end{equation}

In Equation \eqref{eq:nu_svm}, \( \rho \) represents the margin width, which is optimized during training.  \( \nu \in (0,1) \) controls the trade-off between maximizing the margin and penalizing misclassified points. The constraints in Equations \eqref{eq:nu_svm_constraint1} to \eqref{eq:nu_svm_constraint3} ensure that data points are correctly classified while allowing for some margin violations when necessary.

\subsection{Proposed model} Proposed model aims to handle class imbalance and be robust against noise. To handle class imbalance, a function was designed to assign a unique priority to each data point based on its class membership (majority or minority) and its coordinates relative to the decision boundary. Instead of penalizing violations based on their magnitude, the proposed objective function counts the number of violations using a binary variable, making the model more robust against noise. A constraint was introduced to allow the model to permit violations by counting the corresponding misclassifications. The mathematical formulation is as follows:

\begin{equation}
    \min_{\mathbf{w}, b, z} \quad \frac{1}{2} \mathbf{w}^T \mathbf{w} + \sum_{i=1}^{n} c_i z_i
    \label{eq:mixed_integer_svm}
\end{equation}

\noindent \textbf{Subject to:}
\begin{equation}
    y_i (\mathbf{w}^T \mathbf{x}_i + b) \geq 1 - M z_i, \quad \forall i
    \label{eq:mixed_integer_svm_constraint1}
\end{equation}
% \begin{equation}
%     \sum z_i \geq 1
%     \label{eq:mixed_integer_svm_constraint2}
% \end{equation}
\begin{equation}
    z_i \in \{0,1\}
    \label{eq:mixed_integer_svm_constraint3}
\end{equation}

In Equation \eqref{eq:mixed_integer_svm}, the objective function minimizes the norm of the weight vector while incorporating a penalty term controlled by \( c_i \), which assigns a priority to each sample that violates the margin.
 A binary variable \( z_i \) is introduced to quantify margin violations. Unlike the Soft Margin SVM, which penalizes margin violations based on the extent of the violation (Equation \eqref{eq:svm_objective}), Equation \eqref{eq:mixed_integer_svm} penalizes margin violations based on the count of violations. The constraint in Equation~\eqref{eq:mixed_integer_svm_constraint1} ensures that the classification adheres to the SVM formulation.

The priority function used to assign \( c_i \) to each sample is defined as:

\begin{equation}
c_i = \frac{d_i}{|w_{y_i}| }
\label{eq:priority_function}
\end{equation}
\begin{equation}
d_i = \mathbf{w}^T \mathbf{x}_i + b
\label{eq:decision_function}
\end{equation}
Where \( w_{y_i} \), represents the weight assigned to class \( y_i \) based on its membership in the majority or minority class. If accuracy is the desired metric for the user, equal values are assigned across classes. The term \( d_i \) denotes the distance of sample \( i \) from the decision boundary, ensuring that samples closer to the boundary receive higher priority. 

The proposed formulation (Equations \eqref{eq:mixed_integer_svm}–\eqref{eq:mixed_integer_svm_constraint3}) prioritizes the inclusion of samples near the decision boundary. In class imbalance scenarios, minority class samples are given higher priority for inclusion. In noisy scenarios, samples near the decision boundary of each class are given higher priority for inclusion.

Solving this problem (Equations \eqref{eq:mixed_integer_svm}–\eqref{eq:mixed_integer_svm_constraint3}) is challenging due to the presence of both continuous variables $(\mathbf{w}, b)$ and binary variables (\( z_i \)). This problem falls under mixed-integer programming. If the binary variable \( z_i \), the complicating variable, were fixed, the remaining problem would reduce to a hard margin SVM. To tackle this, the Benders decomposition technique \cite{benders2005partitioning} is employed.

The key idea behind Benders decomposition is to avoid solving the original problem with all variables present by decomposing it into a subproblem and a master problem. To establish a connection between the subproblem and the master problem, the dual of the subproblem is solved, and its solution is provided to the master problem. 

The master problem then uses the solution of dual subproblem to add feasibility cut and optimality cuts. These cuts (constraints) make the feasible reagion smaller at each iteration. This iterative exchange of solutions between the master problem and the dual subproblem continues until the feasible region becomes sufficiently small, ultimately leading to the discovery of the optimal solution \cite{rahmaniani2017benders}. Algorithm 1 presents the pseudocode for the proposed Benders decomposition algorithm, while Figure 1 illustrates a high-level overview of the solution procedure.

\begin{algorithm}[ht]
\caption{[Proposed Benders decomposition]} \label{alg:algorithm_name}
\textbf{Input:} Problem data, initial feasible integer solution, tolerance $\epsilon$ \\
\textbf{Output:} Optimal solution or near-optimal solution within tolerance $\epsilon$

\begin{algorithmic}[1]
\State \textbf{Initialization:}
\State \quad Select an initial feasible integer solution.
\State \quad Set lower bound $LB = -\infty$ and upper bound $UB = +\infty$.

\State \textbf{Iterate until convergence:}
\While{$(UB - LB) > \epsilon$}
    \State \textbf{Solve the subproblem:}
    \State \quad Solve the subproblem to obtain either:
    \State \qquad - A feasibility cut (if the subproblem is unbounded), or
    \State \qquad - An optimality cut (if the subproblem is optimal).

    \State \textbf{Update the master problem:}
    \State \quad Add the obtained cut to the master problem.
    \State \quad Refine the solution space based on the cut.

    \State \textbf{Solve the master problem:}
    \State \quad Solve the updated master problem to obtain an improved solution.
    \State \quad Update the lower bound $LB$ with the objective value of the master problem.

    \State \textbf{Update the upper bound:}
    \State \quad Evaluate the objective value of the current solution.
    \State \quad Update the upper bound $UB$ if the current solution improves it.
\EndWhile

\State \textbf{Termination:}
\State \quad Return the optimal or near-optimal solution when $(UB - LB) \leq \epsilon$.
\end{algorithmic}
\end{algorithm}

\begin{figure*}[ht]
    \centering
    \includegraphics[width=0.9\linewidth]{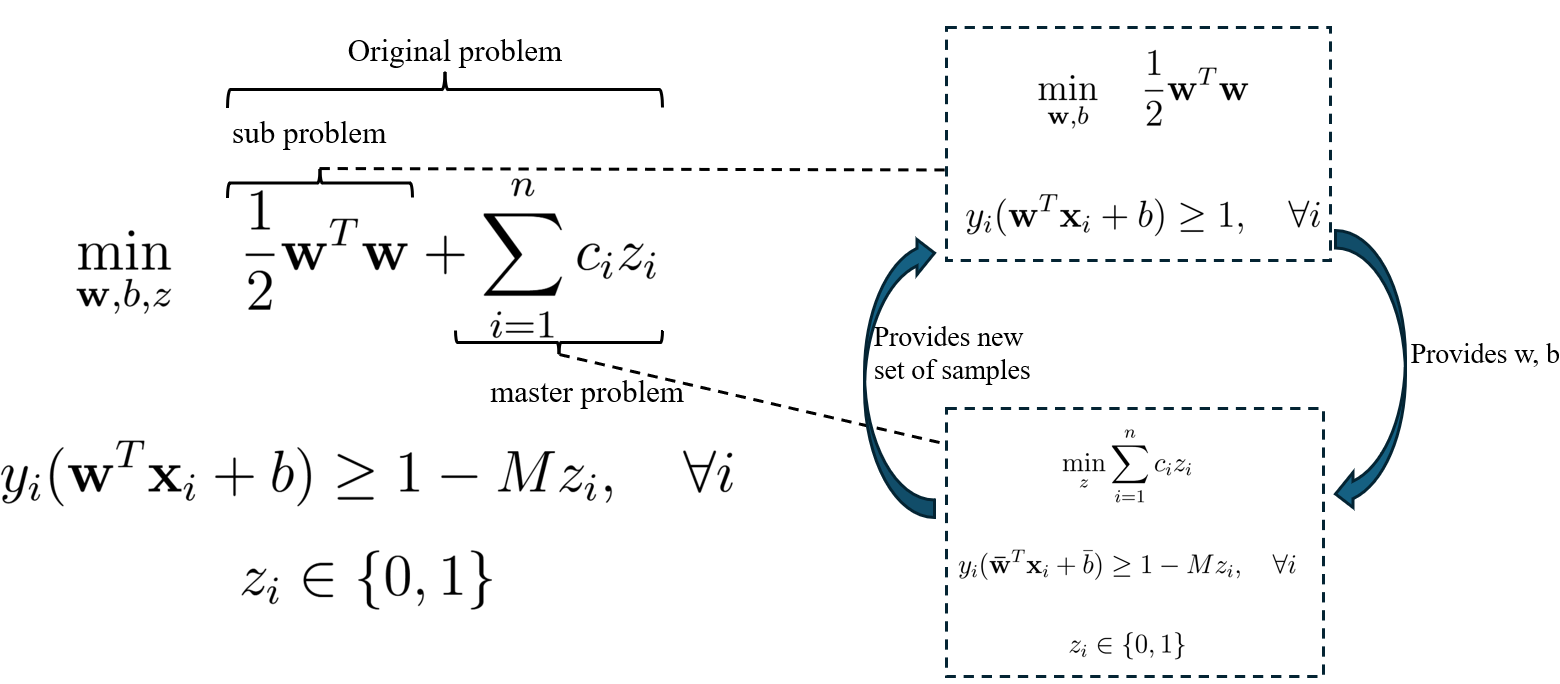}
    \caption{Schematic of the solution procedure for the proposed model using Benders decomposition, which splits the original problem into a subproblem and a master problem. }
    \label{fig:bsvm-framework}
\end{figure*}

The subproblem in the developed model is a hard margin SVM and is presented as follows:

\begin{equation}
    \min_{\mathbf{w}, b} \quad \frac{1}{2} \mathbf{w}^T \mathbf{w}
\end{equation}

\textbf{Subject to:}
\begin{equation}
    y_i (\mathbf{w}^T \mathbf{x}_i + b) \geq 1, \quad \forall i
\end{equation}

The dual form of the subproblem is derived using Lagrange multipliers $\alpha_i$ and is formulated as follows \cite{parand2023basics}:

Minimize:
\begin{equation}
\frac{1}{2} \sum_{i} \sum_{j} \alpha_i \alpha_j y_i y_j x_i^T x_j - \sum_{i} \alpha_i
\end{equation}

Subject to:
\begin{equation}
\sum_{i=1}^{n} \alpha_i y_i = 0, \quad \forall i 
\end{equation}

\begin{equation}
0 \leq \alpha_i, \quad \forall i
\end{equation}

The dual problem's objective function (Equation 16) involves the inner product $x_i^T x_j$. By employing the \textit{kernel trick}, this inner product can be replaced with a kernel function $K(x_i, x_j)$, which implicitly maps the data into a higher-dimensional space. This allows the SVM to handle non-linearly separable data. The kernelized dual problem becomes:

\begin{equation}
\frac{1}{2} \sum_{i} \sum_{j} \alpha_i \alpha_j y_i y_j K(x_i, x_j) - \sum_{i} \alpha_i
\end{equation}

Common choices for $K$ include the polynomial kernel $K(x_i, x_j) = (x_i^T x_j + c)^d$ and the radial basis function (RBF) kernel $K(x_i, x_j) = \exp(-\gamma \|x_i - x_j\|^2)$. The kernel trick avoids explicit computation of coordinates in the high-dimensional space, reducing computational complexity.

The solution for \( \mathbf{w} \) can be formulated using the Lagrange multipliers:

\begin{equation}
    w = \sum_{i=1}^{N} \alpha_i y_i x_i
    \label{eq:w_solution}
\end{equation}

where \( \alpha_i \) are the Lagrange multipliers.

For any support vector \( x_s \), the decision function should satisfy:

\begin{equation}
    y_s \left( \sum_{i=1}^{N} \alpha_i y_i K(x_s, x_i) + b \right) = 1
    \label{eq:support_vector_condition}
\end{equation}

Rearranging to solve for \( b \):

\begin{equation}
    b = y_s - \sum_{i=1}^{N} \alpha_i y_i K(x_s, x_i)
    \label{eq:b_solution}
\end{equation}

Since multiple support vectors exist, \( b \) is typically averaged over all support vectors:

\begin{equation}
    b = \frac{1}{|S|} \sum_{s \in S} \left( y_s - \sum_{i=1}^{N} \alpha_i y_i K(x_s, x_i) \right)
    \label{eq:b_average}
\end{equation}

where S  is the set of support vectors.

The decision function for a new data point \( x \) is given by:

\begin{equation}
    f(x) = \text{sign} \left( \sum_{i=1}^{N} \alpha_i y_i K(x, x_i) + b \right)
    \label{eq:decision_function2}
\end{equation}

Once the values for $\alpha$ are determined, \( \mathbf{w} \) and $b$ can be obtained and incorporated into the master problem. The master problem in the developed model is presented as follows:

\begin{equation}
\min_{z} \sum_{i=1}^{n} c_i z_i
\end{equation}

\textbf{Subject to:}
\begin{equation}
y_i (\mathbf{\bar{w}}^T \mathbf{x}_i + \bar{b}) \geq 1 - M z_i, \quad \forall i
\end{equation}

% \begin{equation}
% \sum z_i \geq 1
% \end{equation}

\begin{equation}
    z_i \in \{0,1\}
\end{equation}

Figure 2 represents the stages of the proposed algorithm in a flowchart. Initially, a feasible solution is created where the classes are perfectly separable, and there are no margin constraint violations. To create the initial feasible solution, an SVM model is first fitted. All samples that are misclassified or fall within the margin are placed in a candidate set called \( h \). The candidate samples and the reduced dataset, where all samples are classified correctly, form the output of the initial solution (Algorithm 2). The reduced dataset and candidate samples are then input into Algorithm 3, which connects the subproblem to the master problem.

The subproblem is a hard margin SVM that is solved, and the decision boundary is updated (Algorithm 4). Subsequently, the master problem assigns priorities to the candidate samples and selects the one with the highest priority for inclusion in the reduced dataset (Algorithm 4). An SVM model is then trained, and the classification performance is evaluated. If misclassification is detected, the next highest-priority sample is selected for addition. This iterative process continues until an appropriate sample is identified for inclusion.

Then, a verification is conducted to determine whether the candidate set is empty. If it is not empty, the updated reduced dataset and candidate samples are returned to Algorithm 3 for further processing. If no samples remain in the candidate set, the model terminates.

\begin{figure}[ht]
    \centering
    \includegraphics[width=0.9\linewidth]{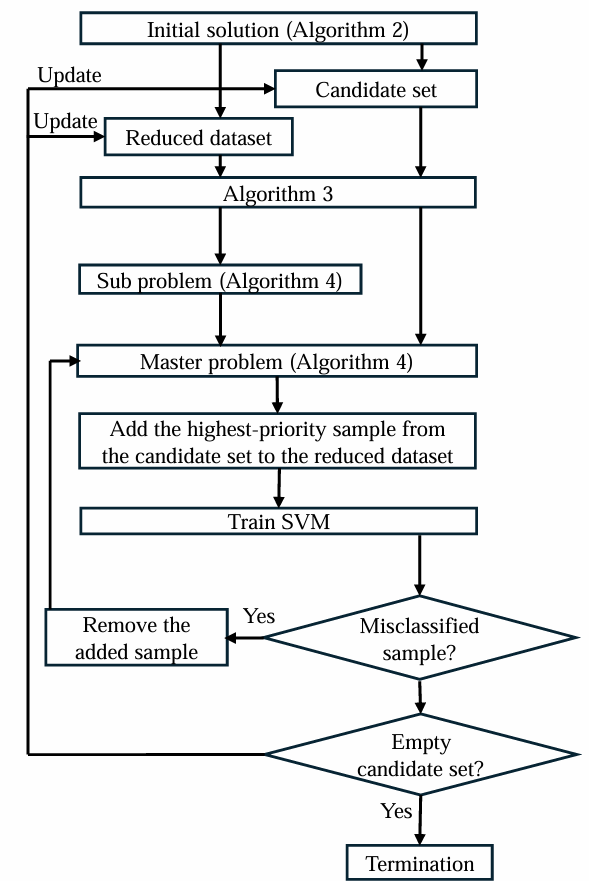}
    \caption{Flowchart of the proposed model}
    \label{fig:enter-label}
\end{figure}

Table 1 introduces the notations, and Algorithms 2, 3, and 4 detail the initialization, the link between the subproblem and the master problem, and the stages in the subproblem and master problem, respectively.

\begin{table}[ht]
    \centering
    \caption{Notations for proposed model}
    \resizebox{\linewidth}{!}{%
    \begin{tabular}{ll}
        \hline
        \textbf{Notation} & \textbf{Description} \\
        \hline
        $X$ & Attributes in initial feasible solution. \\
        $y$ & Labels corresponding to $X$ in initial feasible solution.\\
        $h$ & Samples in original dataset that do not exist in  $X$ (candidate set).\\
        $y_h$ & Labels corresponding to $h$. \\
        $X_{new}$ & Samples in $X$ with an added sample from subproblem. \\
        $y_{new}$ & Labels corresponding to $X_{new}$. \\
        $model$ & SVM model used for classification. \\
        $weights$ & Importance weight for each class label in training. \\
        $b$ & Dictionary of misclassified samples and their computed selection priorities. \\
        $\text{break}$ & Binary flag (0 or 1) indicating whether to stop the iterative process. \\
        \hline
    \end{tabular}}
    \label{tab:notation}
\end{table}

\begin{algorithm}[ht]
\caption{InitialSolution}
\begin{algorithmic}[1]
\Require $X, y, model$
\Ensure $X_{new}, y_{new}, h, y_h$
\State \textbf{Fit} model on $(X, y)$
\State $y_{pred} \gets \text{Predict}(model, X)$
\State $i_2 \gets (y_{pred} \neq y)$ \Comment{Misclassified samples}
\State $d \gets y \times \text{DecisionFunction}(model, X)$
\State $i_1 \gets (d < 1)$ \Comment{Margin constraint violations}
\State $h \gets i_1 \lor i_2$ \Comment{Union of misclassified and violating samples}
\State $y_h \gets \text{Reshape}(y[h], -1, 1)$ \Comment{Reshape to column vector}
\State \textbf{Return} $X[\neg h], y[\neg h], X[h], y_h$
\end{algorithmic}
\end{algorithm}

\begin{algorithm}[ht]
\caption{Connecting master problem and subproblem}
\begin{algorithmic}[1]
\Require $X, y, X_{new}, y_{new}, model, weights$
\State $(X_{new}, y_{new}, h, y_h) \gets \textsc{InitialSolution}(X, y, model)$
\While{$|h| > 0$}
    \State $(X_{new}, y_{new}, h, y_h, break) \gets \textsc{ExtendSamples}(X_{new}, y_{new}, X, y, h, y_h, model, weights)$
    \If{$break = 1$}
        \State \textbf{break}
    \EndIf
\EndWhile
\end{algorithmic}
\end{algorithm}

\begin{algorithm}[ht]
\caption{ExtendSamples}
\begin{algorithmic}[1]
\Require $X_{new}, y_{new}, h, y_h, model, weights$
\State Train $model$ on $(X_{new}, y_{new})$ \Comment{subproblem}
\State $y_{pred} \gets model.\textsc{Predict}(h)$ \Comment{Begin master problem}
\State $correct \gets \emptyset$, $misclassified \gets \emptyset$

\For{$i \gets 1$ to $|h|$}
    \State $margin \gets model.\textsc{DecisionFunction}(h[i])$
    \If{$y_{pred}[i] = y_h[i]$ \textbf{and} $y_h[i] \cdot margin \geq 1$}
        \State $correct \gets correct \cup \{i\}$
    \Else
        \State $misclassified[i] \gets (|margin|) / weights[y_h[i]]$
    \EndIf
\EndFor

\While{$|h| > 0$}
    \State $sorted\_indices \gets \textsc{SortDescending}(misclassified)$
    \For{$idx \in sorted\_indices$}
        \State Add $h[idx]$ and $y_h[idx]$ to $(X_{new}, y_{new})$
        \State Remove $h[idx]$ and $y_h[idx]$ from $(h, y_h)$
        \State Train $model$ on $(X_{new}, y_{new})$
        
        \If{$model.\textsc{Predict}(X_{new}) = y_{new}$}
            \State \Return $(X_{new}, y_{new}, h, y_h, 0)$  \Comment{Optimality cut}
        \EndIf
        
        \State Undo last addition to $(X_{new}, y_{new})$ \Comment{Feasibility cut}
        \State Remove $idx$ from $misclassified$
    \EndFor
    \State \textbf{break}
\EndWhile

\State \Return $(X_{new}, y_{new}, h, y_h, 1)$ \Comment{End master problem}
\end{algorithmic}
\end{algorithm}

\begin{figure*}[ht]
    \centering
    \begin{subfigure}{0.45\textwidth}
        \centering
        \includegraphics[width=\linewidth]{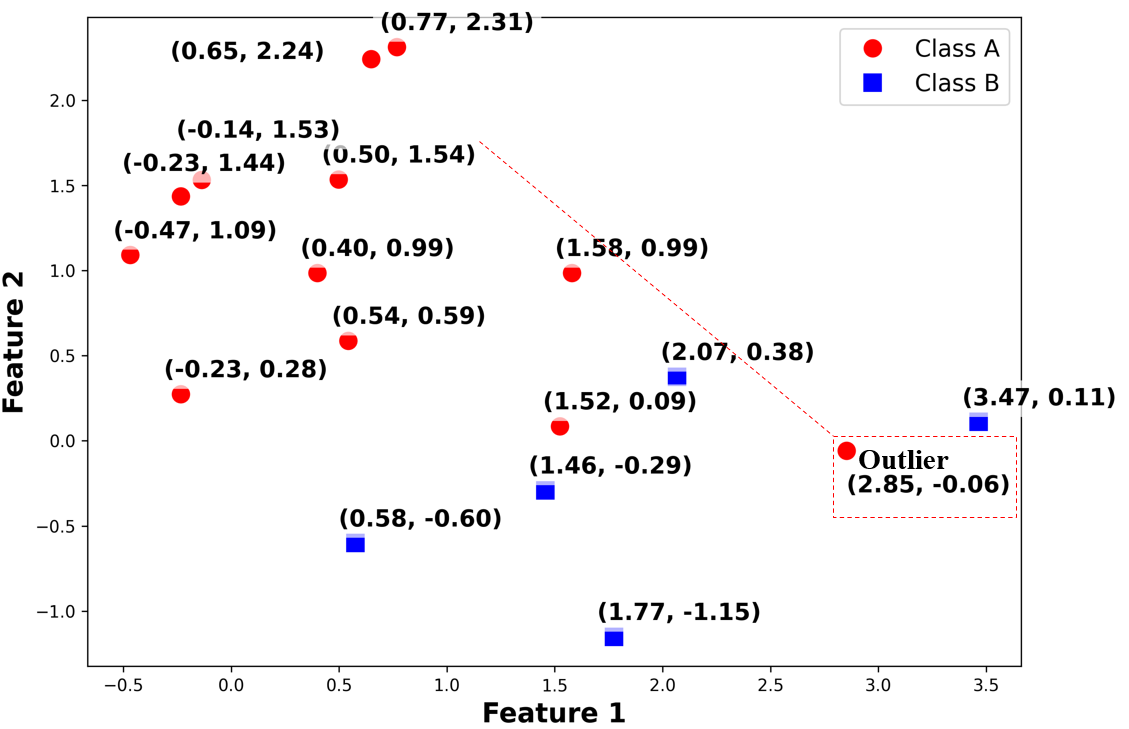}
        \caption{Original data distribution with an imbalance ratio of 2 and the presence of an outlier. Class B represents the minority class and class A represents the majority class.}
        \label{fig:Alpha}
    \end{subfigure}
    \begin{subfigure}{0.40\textwidth}
        \centering
        \includegraphics[width=\linewidth]{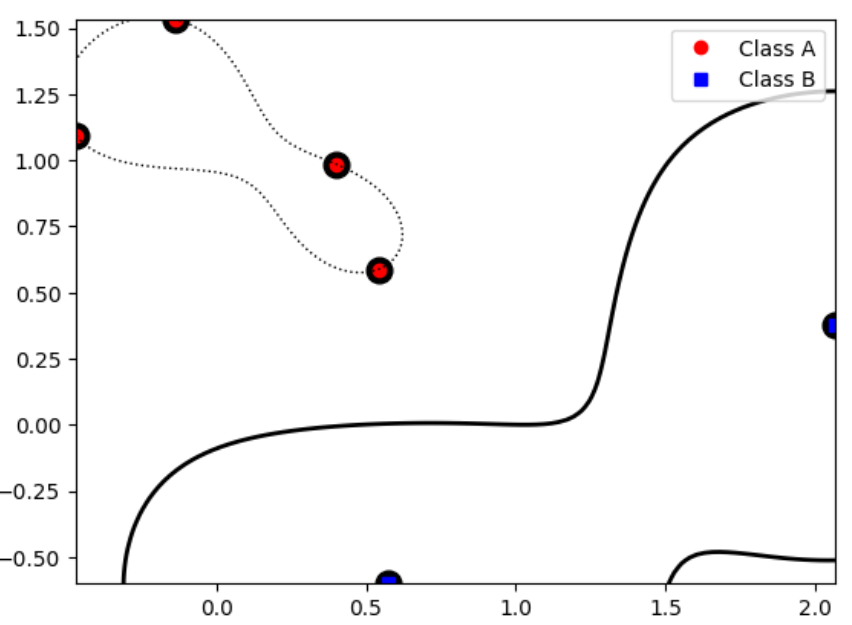}
        \caption{Proposed model initialized with perfectly separable samples. The solid curve is the decision boundary; the dotted curve marks the majority class territory. (Step 0).}
        \label{fig:Bravo}
    \end{subfigure}
    
    \begin{subfigure}{0.40\textwidth}
        \centering
        \includegraphics[width=\linewidth]{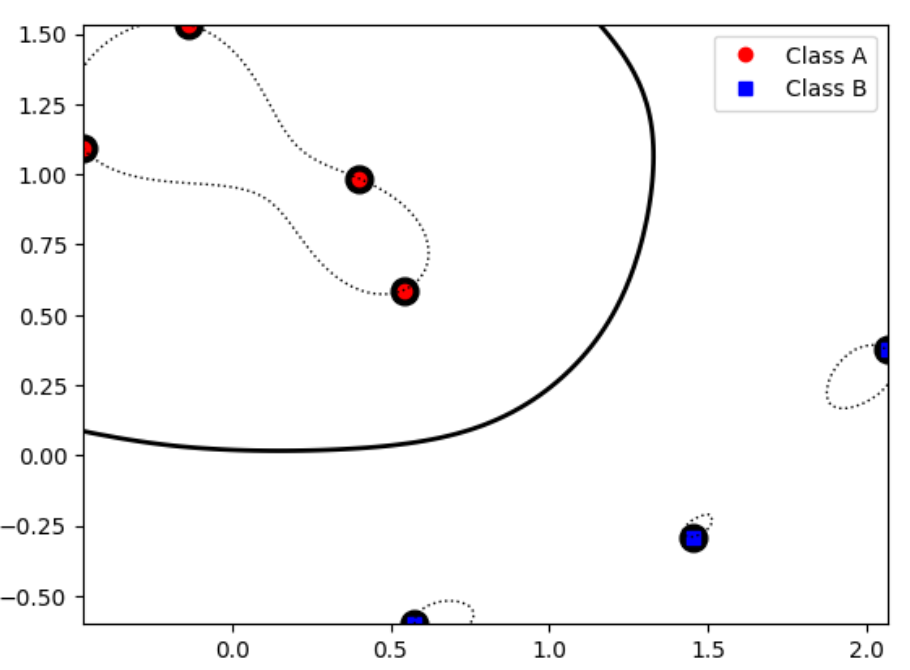}
        \caption{Proposed model with the insertion of a minority class sample at coordinates (1.46, -0.29) as a support vector (Step 1).}
        \label{fig:Charlie}
    \end{subfigure}
    \begin{subfigure}{0.40\textwidth}
        \centering
        \includegraphics[width=\linewidth]{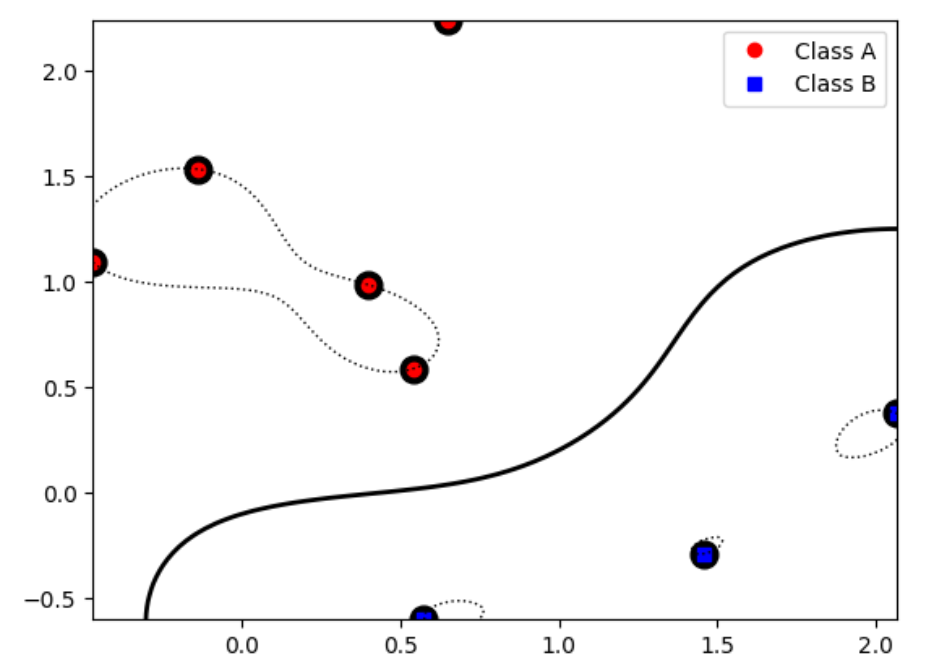}
        \caption{Proposed model with the insertion of a majority class sample at coordinates (0.77, 2.31) as a support vector (Step 2).}
        \label{fig:Delta}
    \end{subfigure}
    
    \begin{subfigure}{0.40\textwidth}
        \centering
        \includegraphics[width=\linewidth]{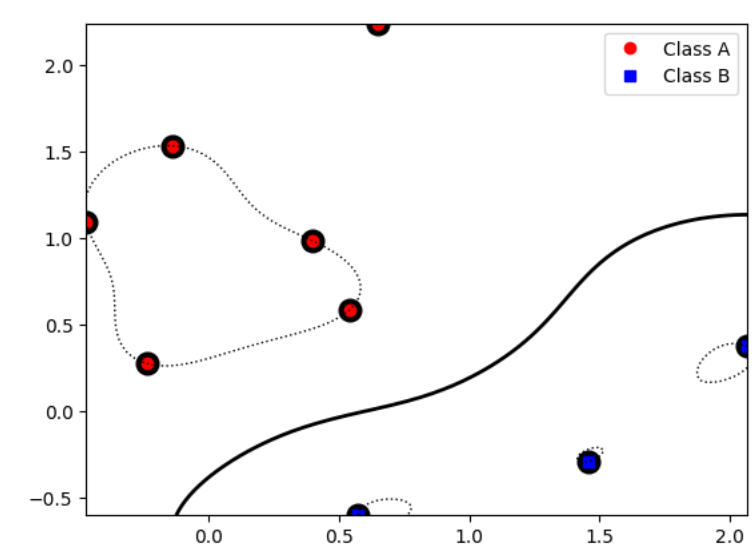}
        \caption{Proposed model with the insertion of a majority class sample at coordinates (-0.23, 0.28) as a support vector. Convergence to optimality with a robust decision boundary favoring the minority class (Step 3).}
        \label{fig:Echo}
    \end{subfigure}
    \begin{subfigure}{0.40\textwidth}
        \centering
        \includegraphics[width=\linewidth]{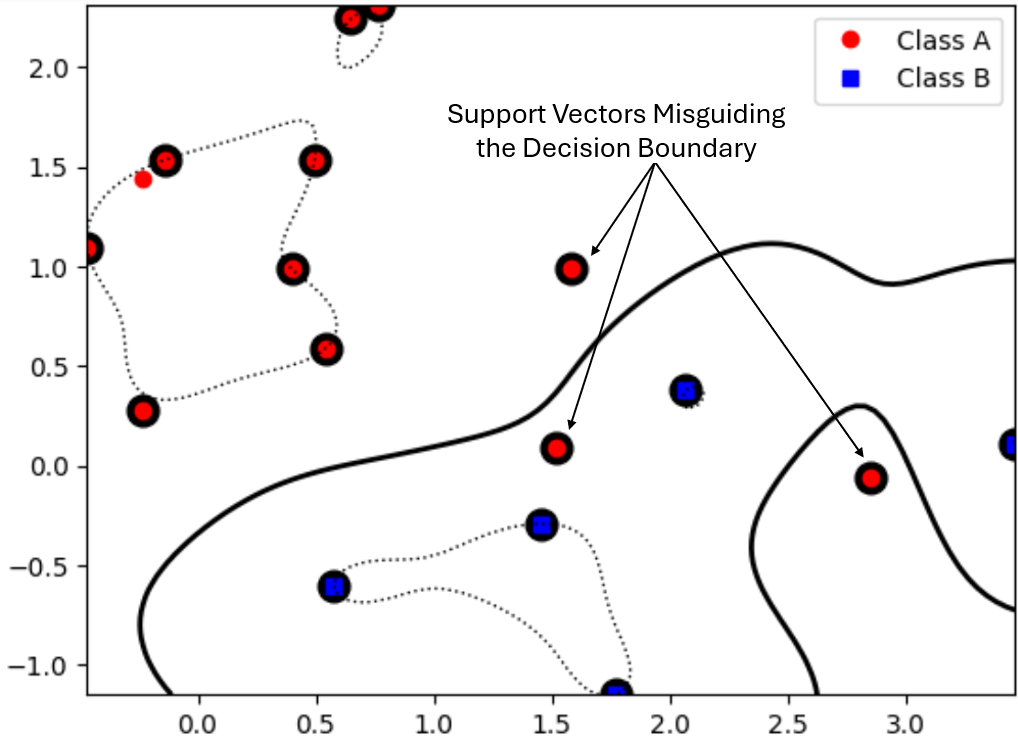}
        \caption{Soft Margin SVM identified support vectors in the presence of all samples and achieved a suboptimal decision boundary.}
        \label{fig:Foxtrot}
    \end{subfigure}
    
    \caption{Comparison of the proposed model's decision boundary evolution and the Soft Margin SVM approach.}
    \label{fig:Comparison}
\end{figure*}

\begin{figure*}[ht]
    \centering
    \begin{subfigure}{0.3\textwidth}
        \centering
        \includegraphics[width=\textwidth]{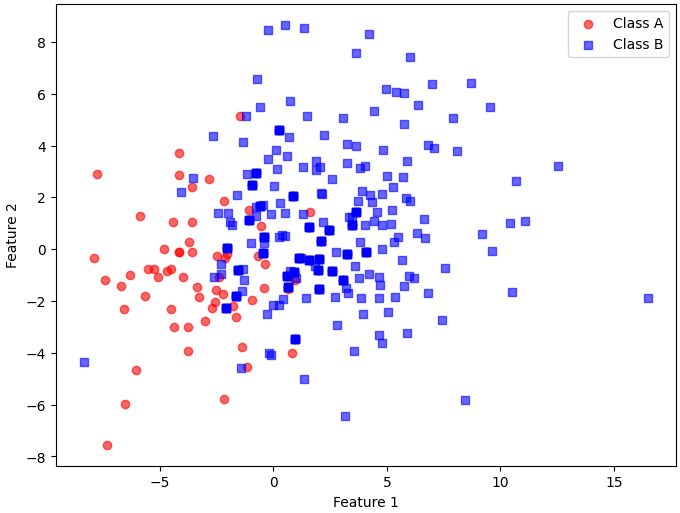}
        \caption{The original data distribution represents a noise scenario with a high overlap between classes.}
    \end{subfigure}
    % \hfill
    \begin{subfigure}{0.3\textwidth}
        \centering
        \includegraphics[width=\textwidth]{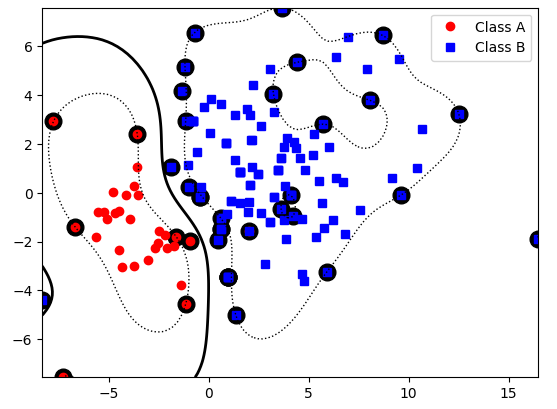}
        \caption{Proposed model gradually updated the support vectors by adding new samples and achieved an accuracy of 72\%.}
    \end{subfigure}
    % \hfill
    \begin{subfigure}{0.3\textwidth}
        \centering
        \includegraphics[width=\textwidth]{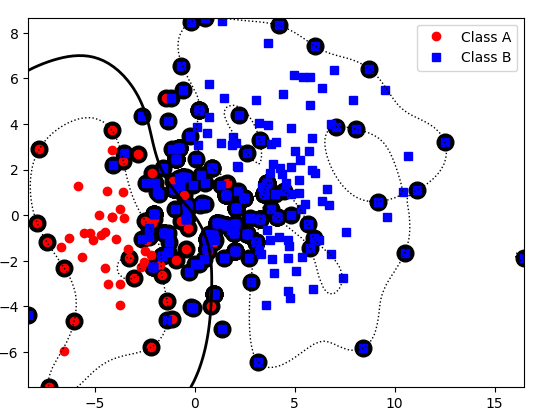}
        \caption{The Soft Margin SVM is trained on the original data representation and achieved an accuracy of 65\%.}
    \end{subfigure}

    \caption{Comparison of data distribution and different model training approaches.}
    \label{noisy_scen}
\end{figure*}

Figure \ref{fig:Comparison} provides a graphical illustration of the steps taken by the proposed model to create a decision boundary for a small synthetic dataset representing a class imbalance scenario (Figure \ref{fig:Alpha}). The decision boundary of the proposed model (Figure \ref{fig:Echo}) is then compared to that of the Soft Margin SVM (Figure \ref{fig:Foxtrot}). The synthetic dataset was intentionally kept small to illustrate the stages of the proposed model. Additionally, a larger synthetic dataset representing a noise scenario was selected. The decision boundaries for this scenario have been visualized to highlight the differences between the proposed model and the Soft Margin SVM, and their performance on unseen data is reported in Figure \ref{noisy_scen}.

\subsection{Analysis}

The primary objective is to assess the effectiveness of the proposed model in handling class imbalance and noisy data scenarios. The proposed model is benchmarked against Soft Margin SVM, Weighted SVM, and NuSVC. The hypotheses are formulated as follows:

\textbf{Hypothesis 1}: For class imbalance data when the minority class is the class of interest for the user, the proposed model achieves higher F1-score for minority class compared to benchmark methods. 

\textbf{Hypothesis 2}: For noisy data when the classes in dataset are equally important, the proposed model achieves higher accuracy compared to benchmark methods.

Additional analyses include reporting the training time, prediction time, number of support vectors, and hyperparameter values for each model.

\subsection{Datasets}
The experimental study utilized a diverse set of binary classification datasets collected from OpenML public repositories. These datasets spanned various domains, including labor contracts, healthcare, finance, Disney movie voice characters, molecular cancer classification, and spacecraft control. Since borderline points play a central role in the proposed model and differentiate it from the Soft Margin SVM, a metric called Fraction of Borderline Points \cite{lorena2019complex} was used to filter the datasets and identify those relevant to this research.

The Fraction of Borderline Points (N1) quantifies dataset complexity by constructing a Minimum Spanning Tree (MST), where each data point is a vertex, and edges are weighted based on pairwise distances. N1 is calculated as the percentage of vertices connected by edges that link samples from different classes. These points typically lie along class boundaries, in overlapping regions, or may represent noisy instances. A higher N1 value suggests a more intricate decision boundary is needed, indicating significant class overlap or complexity in class separation.

\begin{equation}
    N1 = \frac{1}{n} \sum_{i=1}^{n} I\Big( (\mathbf{x}_i, \mathbf{x}_j) \in MST \wedge y_i \neq y_j \Big).
\end{equation}

Datasets with \( N_1 \) greater than 0.05 were selected to ensure their relevance to this study. The characteristics of the chosen datasets for Hypotheses 1 and 2 are presented in Table~\ref{tab:dataset_stats}.

\begin{table*}[ht]
    \centering
    
    \caption{Summary of dataset statistics, including dataset ID, number of instances, number of features, number of missing values, number of instances with missing values, number of numeric and symbolic features, imbalance ratio, and fraction of borderline points. Datasets above the middle horizontal line were used for class imbalance (Hypothesis 1), while those below the line were used for noisy scenarios (Hypothesis 2).}
    \label{tab:dataset_stats}
    
    \begin{tabular}{lp{1cm}p{1cm}p{1cm}p{1cm}p{1cm}p{1cm}p{1cm}p{1cm}p{1cm}}
        \toprule
        \textbf{Dataset Name} & \textbf{Dataset id} & \textbf{ \#\ Inst.} & \textbf{\#\ Feat.} & \textbf{\#\ Miss.} & \textbf{\#\ instance with Miss} & \textbf{\#\ Num. Feat.} & \textbf{\#\ Symb. Feat.} & \textbf{IR} & \textbf{Fraction of Borderline Points} \\
        \midrule
        labor & 4 & 57 & 17 & 326 & 56 & 8 & 9 & 1.80 & 0.08 \\
        ilpd & 1480 & 583 & 11 & 0 & 0 & 9 & 2 & 2.50 & 0.38\\
        credit-approval & 29 & 690 & 16 & 67 & 37 & 6 & 10 & 1.24 & 0.20 \\
        fruitfly & 714 & 125 & 5 & 0 & 0 & 2 & 3 & 1.56 & 0.51 \\
        tecator & 851 & 240 & 125 & 0 & 0 & 124 & 1 & 1.34 & 0.14 \\
        quake & 772 & 2178 & 4 & 0 & 0 & 3 & 1 & 1.24 & 0.47 \\
        students-scores & 43097 & 1000 & 8 & 0 & 0 & 3 & 0 & 1.07 & 0.30 \\
        Titanic & 40704 & 2201 & 4 & 0 & 0 & 3 & 1 & 2.09 & 0.41\\
        \midrule
        tecator & 851 & 240 & 125 & 0 & 0 & 124 & 1 & 1.34 & 0.14\\
        sleuth-case2002 & 902 & 147 & 7 & 0 & 0 & 2 & 5 & 1.12 & 0.37\\
        fruitfly & 714 & 125 & 5 & 0 & 0 & 2 & 3 & 1.56 & 0.51 \\
        leukemia & 1104 & 72 & 7130 & 0 & 0 & 7129 & 1 & 1.85 & 0.18 \\
        cloud & 890 & 108 & 8 & 0 & 0 & 6 & 2 & 2.44 & 0.47 \\
        aids & 346 & 50 & 5 & 0 & 0 & 2 & 3 & 1.00 & 0.70 \\
        prnn-synth & 464 & 250 & 3 & 0 & 0 & 2 & 1 & 1.00 & 0.14 \\
        shuttle-landing-control & 172 & 15 & 7 & 26 & 9 & 0 & 7 & 1.4 & 0.53 \\
        rabe-266 & 782 & 120 & 3 & 0 & 0 & 2 & 1 & 1.00 & 0.08 \\
        \bottomrule
    \end{tabular}
\end{table*}

\subsection{Optimization}

Each dataset was randomly split into training and test sets using an 80/20 ratio with stratified sampling. Within the training data, 80\% was randomly selected for model training, while the remaining 20\% was reserved for hyperparameter tuning (the validation set).

Hyperparameter tuning was performed using a grid search, focusing on the RBF kernel. The RBF kernel was selected for its ability to model complex, non-linear decision boundaries and its applicability to real-world problems. Due to space constraints, the evaluation of other kernels is omitted, as they exhibit similar behavior and lead to comparable conclusions. In the grid search, \( C \) was selected from \{0.1, 1, 10, 100\}, and \( \gamma \) from \{1, 0.1, 0.01, 0.001\} for Soft Margin SVM, Weighted SVM, and the proposed model. For NuSVC, the hyperparameter \( \nu \) was optimized instead of \( C \), with values chosen from \{0.1, 0.75, 1\}.

In class imbalance scenarios, a weighting method \cite{king2001logistic} was utilized to assign higher weights to the minority class, with the weights (\(w_i\)) defined as:

\begin{equation}
w_i = \frac{n}{k n_i}
\end{equation}

where \( n \) is the total number of samples, \( n_i \) is the number of samples in each class, and \( k \) is the number of classes (in this case, \( k = 2 \)), which determines the weight \( w_i \).

The F1-score of the minority class (positive class) and accuracy were used as optimization metrics in class imbalance and noise scenarios. The top-performing model was subsequently evaluated on the holdout test data, and performance was reported for each method.

Statistical tests for hypothesis evaluation were conducted using performance scores (F1-score for class imbalance and accuracy for noisy data) across datasets. A Wilcoxon signed-rank test \cite{rainio2024evaluation} was performed to compare the scores of different methods (e.g., the proposed model versus Soft Margin SVM) across datasets. Statistical significance was determined at \( p < 0.05 \).

The experiments were conducted on a high-performance computing (HPC) cluster managed by SLURM. Each job was allocated 2 CPU cores and 24 GiB of RAM and executed on a single node. The experiments were performed using Python 3.8.6 (GCCcore 10.2.0).

\section{Result}

The proposed model's performance was assessed in comparison to Soft Margin SVM, Weighted SVM, and NuSVC across multiple datasets.

\textbf{Hypothesis 1:} Table~\ref{tab:svm_comparison1} presents the F1-scores for the minority and majority classes in both the training and testing phases. In the training phase, the proposed model underperformed or achieved similar results to the benchmark methods. However, in the testing phase, the proposed model consistently outperformed the benchmark methods. The difference in performance was considerable in specific datasets. For instance, in the \textit{fruitfly} dataset, the proposed model achieved an F1-score of 62\% for the minority class, compared to 35\% for Soft Margin and Weighted SVM and 58\% for NuSVC. Similarly, in the \textit{ilpd} dataset, the proposed model outperformed all methods with an F1-score of 56\% for the minority class, whereas Soft Margin SVM and Weighted SVM achieved 51\% and NuSVC achieved 50\%, respectively. The performance of Weighted SVM was similar to Soft Margin SVM, with the exception of the \textit{quake} dataset. 

\begin{figure*}[ht]
    \centering
    \includegraphics[width=0.49\linewidth]{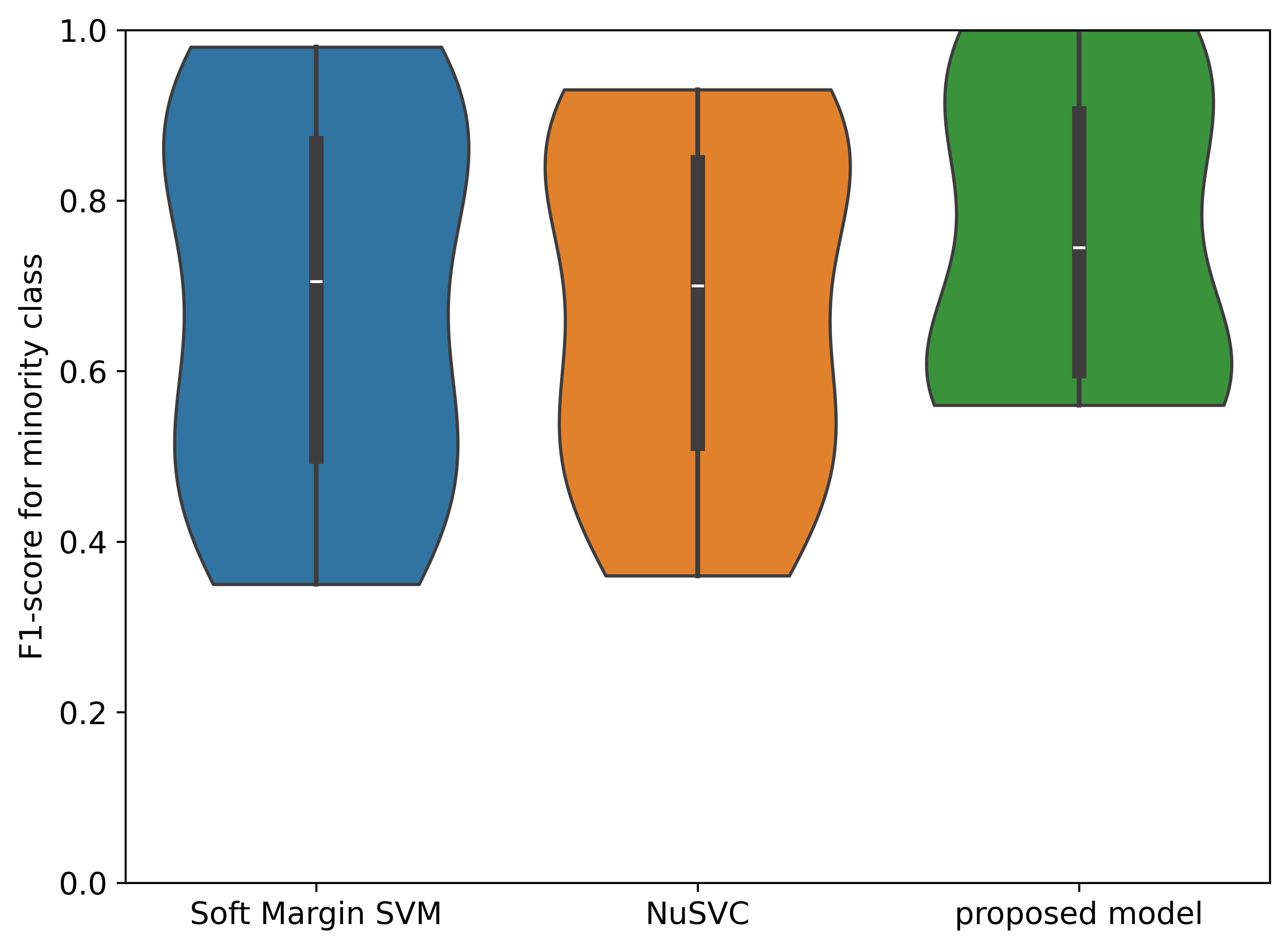}
    \includegraphics[width=0.49\linewidth]{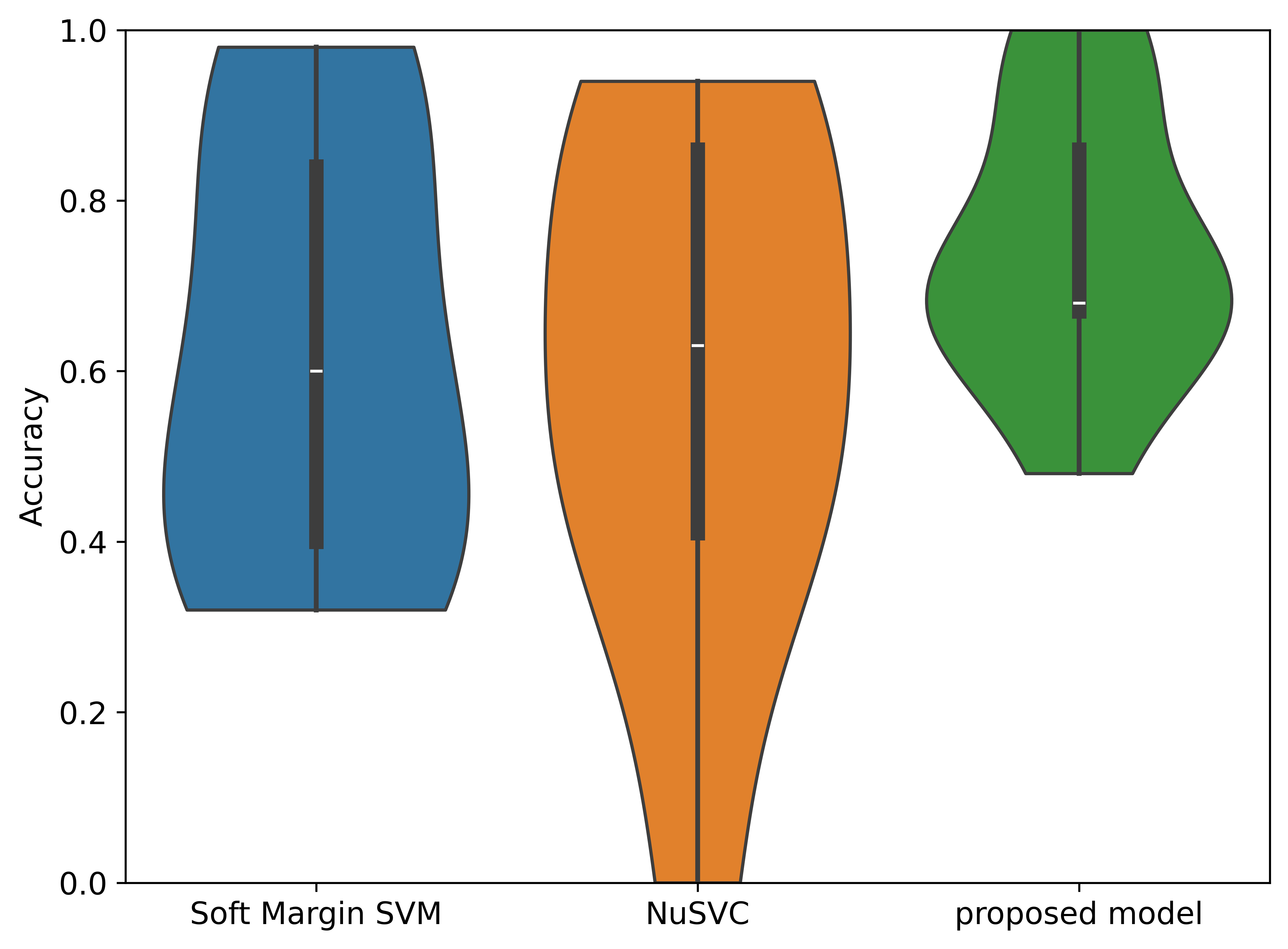}
    \caption{Comparing F1-score (left) and accuracy (right) of Soft Margin SVM, NuSVC, and the proposed model across datasets.}
    \label{fig:comparison_plots}
\end{figure*}

The Wilcoxon test revealed a statistically significant difference (\( p < 0.05 \)) between the performance of the proposed model and the benchmark methods when evaluated across eight different datasets.
 Figure~\ref{fig:comparison_plots} (left) illustrates the distribution of F1-scores for the minority class across different models. The proposed model shows a higher F1-score distribution compared to Soft Margin SVM and NuSVC. As Weighted SVM demonstrated similar behavior to Soft Margin SVM, it was excluded from the illustration in Figure 6 (left) to enhance clarity and conciseness.

\textbf{Hypothesis 2:} Table~\ref{tab:svm_comparison} presents accuracy (the evaluation metric for the hypothesis) as well as the macro-average precision, recall, and F1-score. The proposed model consistently achieved higher accuracy across all datasets compared to Soft Margin SVM in both the training and test phases. Notably, on the \textit{leukemia} dataset, the proposed model achieved an accuracy of 67\%, surpassing Soft Margin SVM by up to 34\%. Similarly, on the \textit{aids} dataset, the proposed model attained 70\% accuracy, reflecting a 20\% improvement over NuSVC.

The Wilcoxon test indicates a significant difference between the proposed model's performance and the benchmark methods  (\( p < 0.05 \)) across nine datasets. Figure~\ref{fig:comparison_plots} (right) represents the accuracy distribution across models, where the proposed model maintains higher accuracy compared to Soft Margin SVM and NuSVC. When accuracy is the primary evaluation metric, equal class weights are assigned to both classes, making Weighted SVM redundant and thus excluded from the analysis.

Table~V compares the training time, prediction time, and the number of support vectors (\#SV) across Soft Margin SVM, NuSVC, and the proposed model. While the training time for the proposed model is higher compared to the benchmark methods, the prediction time remains competitive due to the lower number of support vectors.

\begin{figure*}[ht]
    \centering
    \includegraphics[width=1\linewidth]{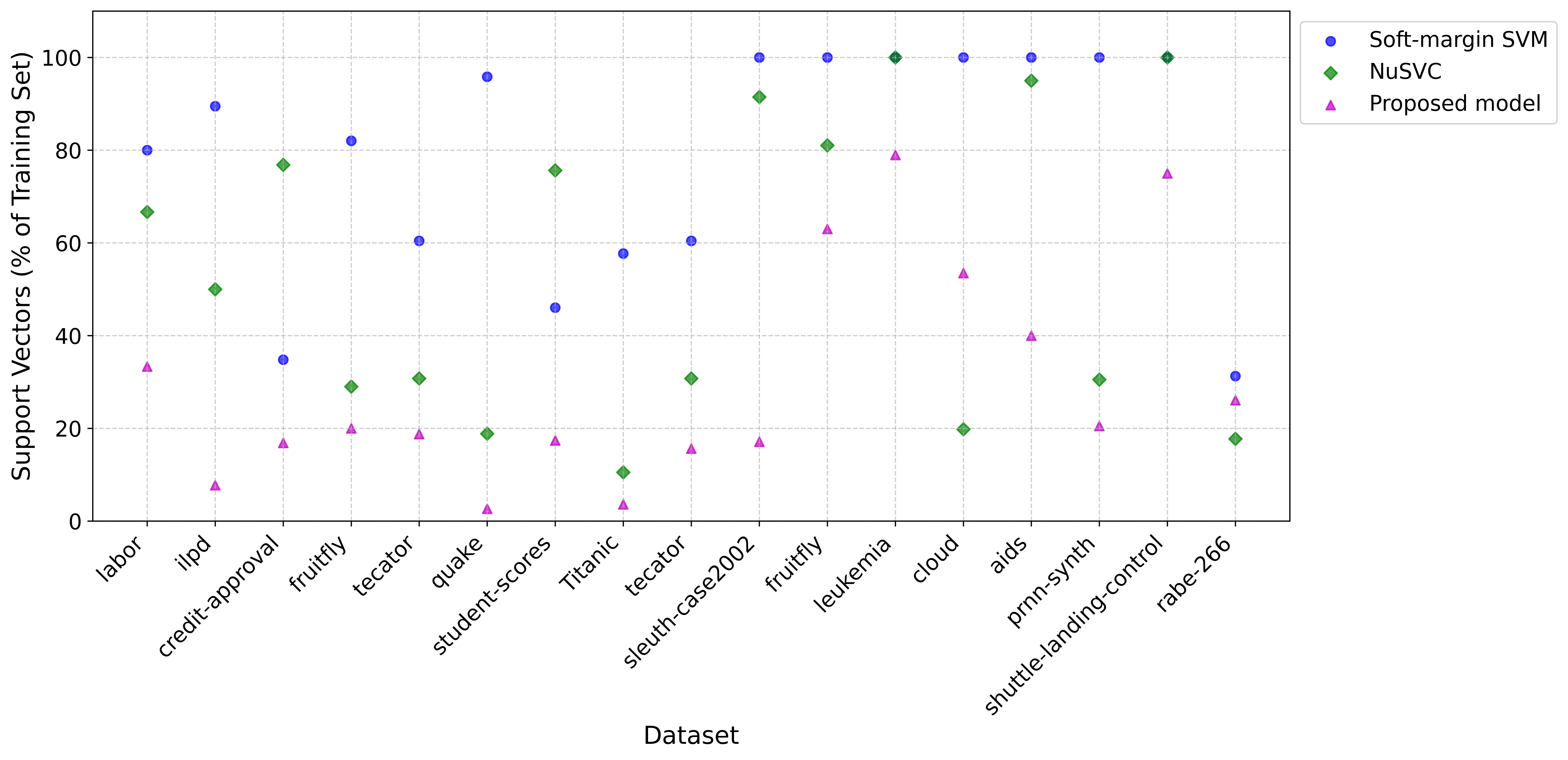}
    \caption{Support vectors as a percentage of the training set across different models.}
    \label{fig:SVCount}
\end{figure*}

Figure \ref{fig:SVCount} compares the percentage of support vectors across datasets. Soft Margin  SVM (blue) selects the most support vectors, often near 100\%. NuSVC (green) uses fewer but remains relatively high. The proposed model (pink) consistently selects the fewest number of support vectors.

Figure 8 presents the hyperparameter tuning results for models on the quake dataset, optimized for F1-score. The quake dataset was selected because it was the only dataset where Soft Margin SVM and Weighted SVM achieved different optimal values. Each subplot illustrates the model's performance across varying hyperparameter values, with the highest F1-score indicated by a red marker.

\begin{table*}[ht]
    \centering
    \caption{Comparison of different SVM models across eight datasets for class imbalance (hypothesis 1). The highest achieved F1-score for minority class is highlighted in bold.}
    \begin{tabular}{l lcccccccc}
        \toprule
        \multirow{3}{*}{Dataset} & \multirow{3}{*}{Class} & \multicolumn{2}{c}{Soft Margin SVM} & \multicolumn{2}{c}{Weighted SVM} & \multicolumn{2}{c}{NuSVC} & \multicolumn{2}{c}{Proposed Model} \\
        \cmidrule(lr){3-4} \cmidrule(lr){5-6} \cmidrule(lr){7-8} \cmidrule(lr){9-10}
        & & Train & Test & Train & Test & Train & Test & Train & Test \\
        \midrule
        \multirow{2}{*}{labor} & Minority & 0.97 & 0.89 & 0.97 & 0.89 & 1.00 & 089 & 0.97 & \textbf{1.00}\\
                                    & Majority & 0.98 & 0.93 & 0.98 & 0.93 & 1.00 & 0.93 & 0.98 & 1.00 \\
        \midrule
        \multirow{2}{*}{ilpd} & Minority & 0.59 & 0.51 & 0.59 & 0.51& 0.64 & 0.50 & 0.57 & \textbf{0.56} \\
                                    & Majority & 0.65 & 0.55 & 0.65 & 0.55 & 0.85 & 0.74 & 0.59 & 0.55 \\
        \midrule
        \multirow{2}{*}{credit-approval} & Minority & 0.92 & 0.84 & 0.92 & 0.84 & 0.85 & 0.82 & 0.85 & \textbf{0.87} \\
                                    & Majority & 0.94 & 0.87 & 0.94 & 0.87 & 0.88 & 0.84 & 0.86 & \ 0.87 \\
        \midrule
        \multirow{2}{*}{fruitfly} & Minority & 0.69 & 0.35 & 0.69 & 0.35 &  0.35 & 0.58 & 0.58 & \textbf{0.62}\\
                                    & Majority & 0.78 & 0.44 & 0.78 & 0.44 & 0.51 & 0.62 & 0.49 & 0.58 \\
        \midrule
        \multirow{2}{*}{tecator} & Minority & 0.94 & 0.98 & 0.94 & 0.98 & 0.99 & 0.93 & 0.96 & \textbf{1.00}\\
                                    & Majority & 0.96 & 0.98 & 0.96 & 0.98 & 1.00 & 0.95 & 0.97 & 1.00 \\
        \midrule
        \multirow{2}{*}{quake} & Minority & 0.54 & 0.47 & 0.15 & 0.11 & 0.53 & 0.52 & 0.61 & \textbf{0.61}\\
                                    & Majority & 0.58 & 0.54 & 0.70 & 0.70& 0.40 & 0.38 & 0.12 & 0.06 \\
        \midrule
        \multirow{2}{*}{students-scores} & Minority & 0.92 & 0.86 &  0.92 & 0.86 & 0.89 & 0.83 & 0.91 & \textbf{0.87}\\
                                    & Majority & 0.93 & 0.87 &  0.93 & 0.87  & 0.90 & 0.85 & 0.92 & 0.89\\
        \midrule
        \multirow{2}{*}{ Titanic} & Minority & 0.62 & \textbf{0.57} & 0.62 & 0.57 & 0.37 & 0.36 & 0.61 & \textbf{0.57}\\
                                    & Majority & 0.83 & 0.80 & 0.83 & 0.80 & 0.59 & 0.60 & 0.82 & 0.79\\
        % Add more rows as needed
        \bottomrule
    \end{tabular}
    \label{tab:svm_comparison1}
\end{table*}

\begin{table*}[ht]
    \centering
    \caption{Comparison of different SVM models across datasets for hypothesis 2. The highest achieved Accuracy is highlighted in bold. Macro-average recall, precision, and F1-score are also reported in the table.}
    \resizebox{\linewidth}{!}{%
    \begin{tabular}{p{3.8cm} lcccc cccc cccc}
        \toprule
        \multirow{3}{*}{Dataset} & \multirow{3}{*}{Set} 
        & \multicolumn{4}{c}{Soft Margin SVM} 
        & \multicolumn{4}{c}{NuSVC} 
        & \multicolumn{4}{c}{Proposed Model} \\
        \cmidrule(lr){3-6} \cmidrule(lr){7-10} \cmidrule(lr){11-14}
        & & Accuracy & Recall & Precision & F1-score 
          & Accuracy & Recall & Precision & F1-score 
          & Accuracy & Recall & Precision & F1-score \\
        \midrule
        \multirow{2}{*}{tecator} 
        & Train & 0.95 & 0.96 & 0.95 & 0.95  & 0.99 & 0.99 & 1.00 & 0.99  & 0.95 & 0.95 & 0.95 & 0.95 \\
        & Test  &0.98 & 0.98 & 0.98 & 0.98 & 0.94 & 0.93 & 0.94 & 0.94  & \textbf{1.00} & 1.00 & 1.00 & 1.00 \\
        \midrule
        \multirow{2}{*}{sleuth-case2002} 
        & Train & 0.47 & 0.24 & 0.50 & 0.32  & 0.82 & 0.82 & 0.82 & 0.82  & 0.68 & 0.68 & 0.68 & 0.67 \\
        & Test  & 0.47 & 0.23 & 0.50 & 0.32  & 0.63 & 0.63 & 0.62 & 0.62  & \textbf{0.67} & 0.68 & 0.67 & 0.67 \\
        \midrule
        \multirow{2}{*}{fruitfly} 
        & Train & 0.39 & 0.20 & 0.50 & 0.28  & 0.73 & 0.72 & 0.70 & 0.70  & 0.71 & 0.75 & 0.64 & 0.64 \\
        & Test  & 0.40 & 0.20 & 0.50 & 0.29  & 0.36 & 0.33 & 0.33 & 0.33  & \textbf{0.48} & 0.42 & 0.43 & 0.42 \\
        \midrule
        \multirow{2}{*}{leukemia} 
        & Train & 0.35 & 0.18 & 0.50 & 0.26  & 1.00 & 1.00 & 1.00 & 1.00  & 0.93 & 0.95 & 0.90 & 0.92 \\
        & Test  & 0.33 & 0.17 & 0.50 & 0.25  & \textbf{0.67} & 0.33 & 0.50 & 0.40  & \textbf{0.67} & 0.33 & 0.50 & 0.40 \\
        \midrule
        \multirow{2}{*}{cloud} 
        & Train & 0.29 & 0.15 & 0.50 & 0.23  & 0.56 & 0.54 & 0.55 & 0.53  & 0.71 & 0.35 & 0.50 & 0.41 \\
        & Test  & 0.32 & 0.16 &  0.50 & 0.24  & 0.41 & 0.49 & 0.49 & 0.41  & \textbf{0.68} & 0.34 & 0.50 & 0.41 \\
        \midrule
        \multirow{2}{*}{aids} 
        & Train & 0.62 & 0.79 & 0.62 & 0.56  & 0.72 & 0.73 & 0.72 & 0.72  & 0.62 & 0.63 & 0.62 & 0.62 \\
        & Test  & 0.60 & 0.78 & 0.60 & 0.52  & 0.50 & 0.50 & 0.50 & 0.45  & \textbf{0.70} & 0.71 & 0.70 & 0.70 \\
         \midrule
         \multirow{2}{*}{prnn-synth} 
        & Train & 0.81 & 0.84 & 0.81 & 0.81  & 0.83 & 0.84 & 0.83 & 0.83  & 0.85 & 0.85 & 0.85 & 0.85 \\
        & Test  & 0.84 & 0.86& 0.86 & 0.84 & \textbf{0.86} & 0.86 &0.86 & 0.86  & \textbf{0.86} & 0.86 & 0.86 & 0.86 \\
        \midrule
         \multirow{2}{*}{shuttle-landing-control} 
        & Train & 0.58 & 0.29 &   0.50& 0.37 &  1.00 & 1.00&1.00 & 1.00  & 0.58 & 0.29 & 0.50 & 0.37 \\
        & Test  & \textbf{0.67} & 0.33 & 0.50 & 0.40  & 0.00 &  0.00 &  0.00 &  0.00  & \textbf{0.67} & 0.33 & 0.50 & 0.40 \\
        \midrule
         \multirow{2}{*}{ rabe-266} 
        & Train & 0.99 & 0.99 & 0.99 & 0.99  & 0.99 & 0.99& 0.99& 0.99  & 0.97 & 0.97 & 0.97 & 0.97 \\
        & Test  & \textbf{0.96} & 0.96 & 0.96 & 0.96 & 0.92 & 0.92 & 0.92 & 0.92  & \textbf{0.96} & 0.96 & 0.96 & 0.96 \\
        \bottomrule
    \end{tabular}}
    \label{tab:svm_comparison}
\end{table*}

\begin{table*}[ht]
    \centering
    \caption{Training and prediction times with support vector counts for different models across datasets.}
    \label{tab:train_pred_times}
    \sisetup{scientific-notation = true, round-mode = places, round-precision = 2}
    \resizebox{\textwidth}{!}{%
    \begin{tabular}{llllllllll}
        \toprule
        \textbf{Dataset} & \multicolumn{3}{c}{\textbf{Soft Margin  SVM}} & \multicolumn{3}{c}{\textbf{NuSVC}} & \multicolumn{3}{c}{\textbf{Proposed Model}} \\
        \cmidrule(lr){2-4} \cmidrule(lr){5-7} \cmidrule(lr){8-10}
        & \textbf{Train Time} & \textbf{Pred Time} & \textbf{\#SV} & \textbf{Train Time} & \textbf{Pred Time} & \textbf{\#SV} & \textbf{Train Time} & \textbf{Pred Time} & \textbf{\#SV} \\
        \midrule
        labor & \num{1e-3}  & \num{1.24e-5} & 36 & \num{1e-3} & \num{1.22e-5} & 30 & \num{7e-2} & \num{1.09e-5} & 15 \\
        ilpd & \num{1e-2} & \num{2.43e-5} & 417 & \num{1e-2} & \num{1.4e-5} & 233 & \num{3.54} & \num{3e-6} & 36 \\
        credit-approval & \num{1e-2} & \num{1.68e-5} & 192 & \num{2e-2} & \num{3.76e-5} & 424 & \num{1.58} & \num{1.02e-5} & 93 \\
        fruitfly & \num{1e-3} & \num{9.3e-6} & 82 & \num{1e-3} & \num{6e-6} & 29 & \num{2.2e-1} & \num{5.7e-6} & 20 \\
        tecator & \num{1e-3} & \num{1.7e-5} & 116 & \num{1e-3} & \num{9.9e-6} & 59 & \num{5.5e-1} & \num{7e-6} & 36 \\
        quake & \num{1.6e-1} & \num{8.63e-5} & 1669 & \num{4e-2} & \num{1.56e-5} & 328 & \num{4.602e1} & \num{2.4e-6} & 46 \\
        student-scores & \num{2e-2} & \num{2.23e-5} & 368 & \num{3e-2} & \num{3.23e-5} & 605 & \num{5.16} & \num{8.5e-6} & 139 \\
        Titanic & \num{8e-2} & \num{4.83e-5} & 1015 & \num{2e-2} & \num{8.7e-6} & 185 & \num{4.019e1} & \num{2.9e-6} & 63 \\
        \midrule
        tecator & \num{1e-3}  & \num{1.6e-5} & 116 & \num{1e-3} & \num{9.4e-6} & 59 & \num{7.4e-1} & \num{6.1e-6} & 30 \\
        sleuth-case2002 & \num{1e-3} & \num{9.8e-6} & 117 & \num{1e-3} & \num{9.7e-6} & 107 & \num{3.8e-1} & \num{5e-6} & 20 \\
        fruitfly & \num{1e-3} & \num{1.03e-5} & 100 & \num{1e-3} & \num{8.7e-6} & 81 & \num{3.1e-1} & \num{7.9e-6} & 63 \\
        leukemia & \num{1e-2} & \num{2.238e-4} & 57 & \num{1e-2} & \num{2.248e-4} & 57 & \num{5.6e-1} & \num{1.834e-4} & 45 \\
        cloud & \num{1e-3} & \num{1.01e-5} & 86 & \num{1e-3} & \num{6.2e-6} & 17 & \num{1.1e-1} & \num{7.6e-6} & 46 \\
        aids & \num{1e-3} & \num{1.44e-5} & 40 & \num{1e-3} & \num{1.38e-5} & 38 & \num{4e-2} & \num{1.34e-5} & 16 \\
        prnn-synth & \num{1e-3} & \num{1.22e-5} & 200 & \num{1e-3} & \num{5.6e-6} & 61 & \num{1.1} & \num{4.3e-6} & 41 \\
        shuttle-landing-control & \num{1e-3} & \num{4.19e-5} & 12 & \num{1e-3} & \num{4.12e-5} & 12 & \num{1e-2} & \num{4.2e-5} & 9 \\
        rabe-266 & \num{1e-3} & \num{6.5e-6} & 30 & \num{1e-3} & \num{5.9e-6} & 17 & \num{7e-2} & \num{6.2e-6} & 25 \\
        \bottomrule
    \end{tabular}}
\end{table*}

\begin{figure*}[ht]
    \centering
    % Top row
    \begin{subcaptionblock}{.45\textwidth}
        \centering
        \includegraphics[width=\textwidth]{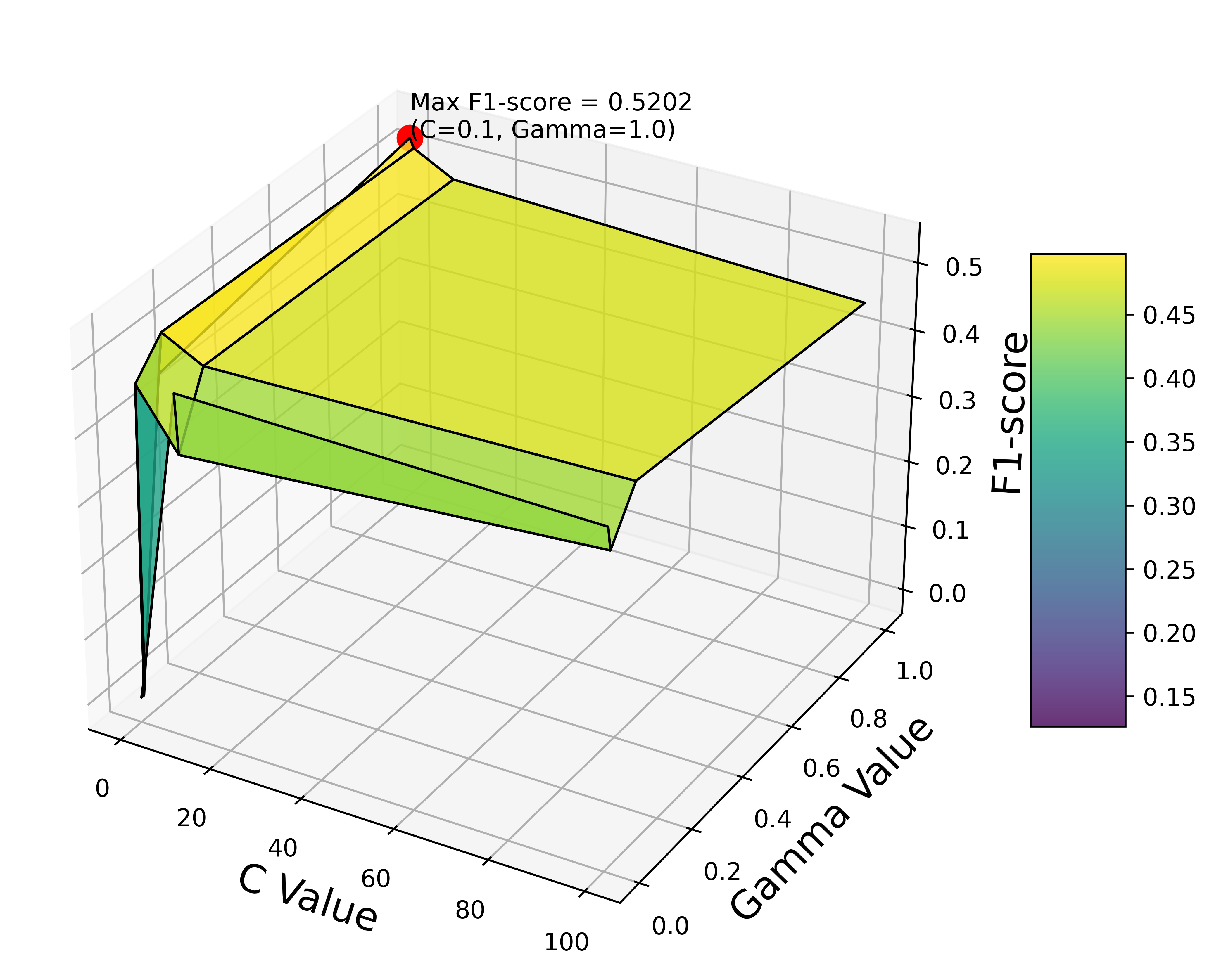}
        \caption{Soft Margin SVM}
    \end{subcaptionblock}
    \begin{subcaptionblock}{.45\textwidth}
        \centering
        \includegraphics[width=\textwidth]{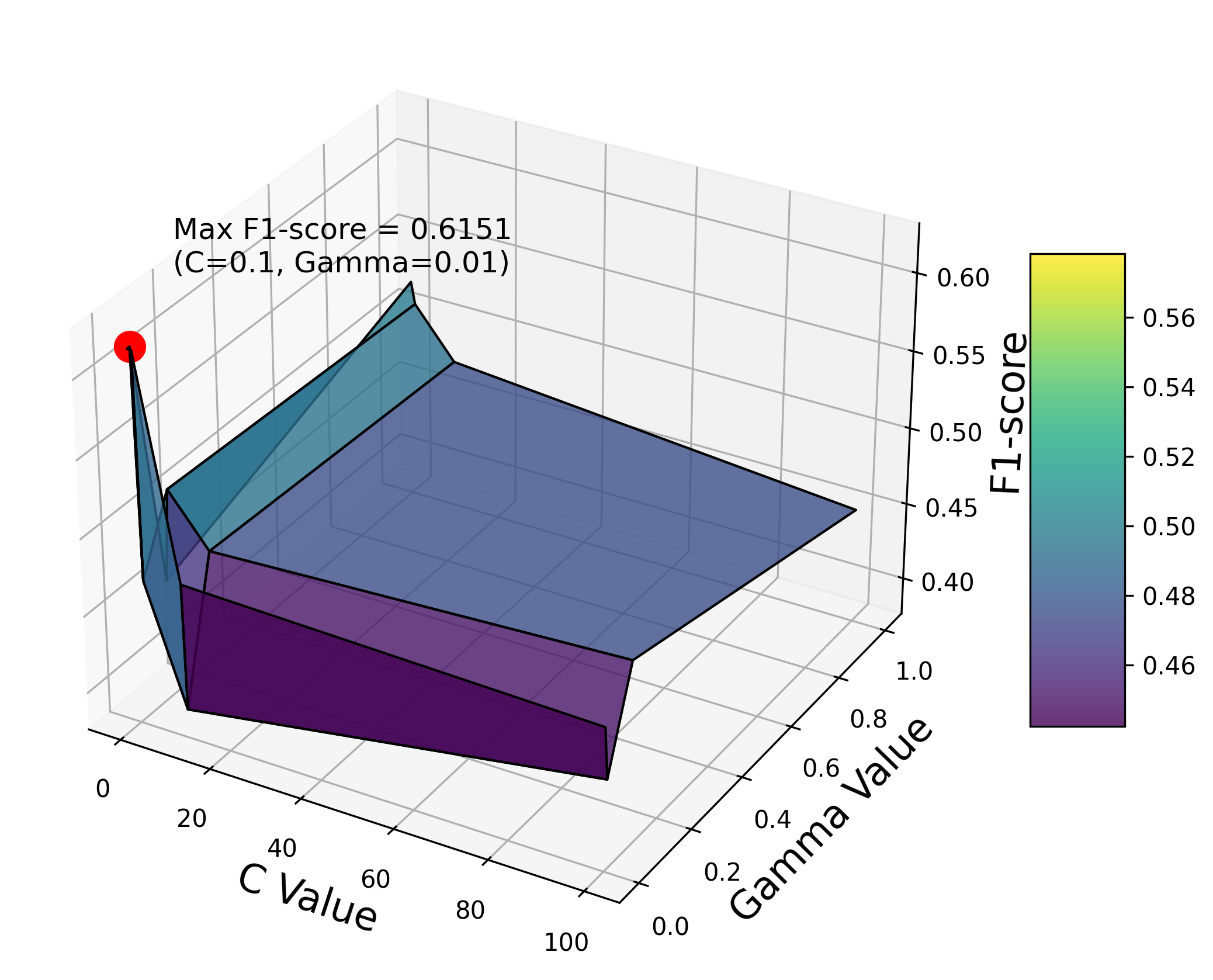}
        \caption{Weighted SVM}
    \end{subcaptionblock}

    % Add vertical space between rows
    \vspace{1em}

    % Bottom row
    \begin{subcaptionblock}{.45\textwidth}
        \centering
        \includegraphics[width=\textwidth]{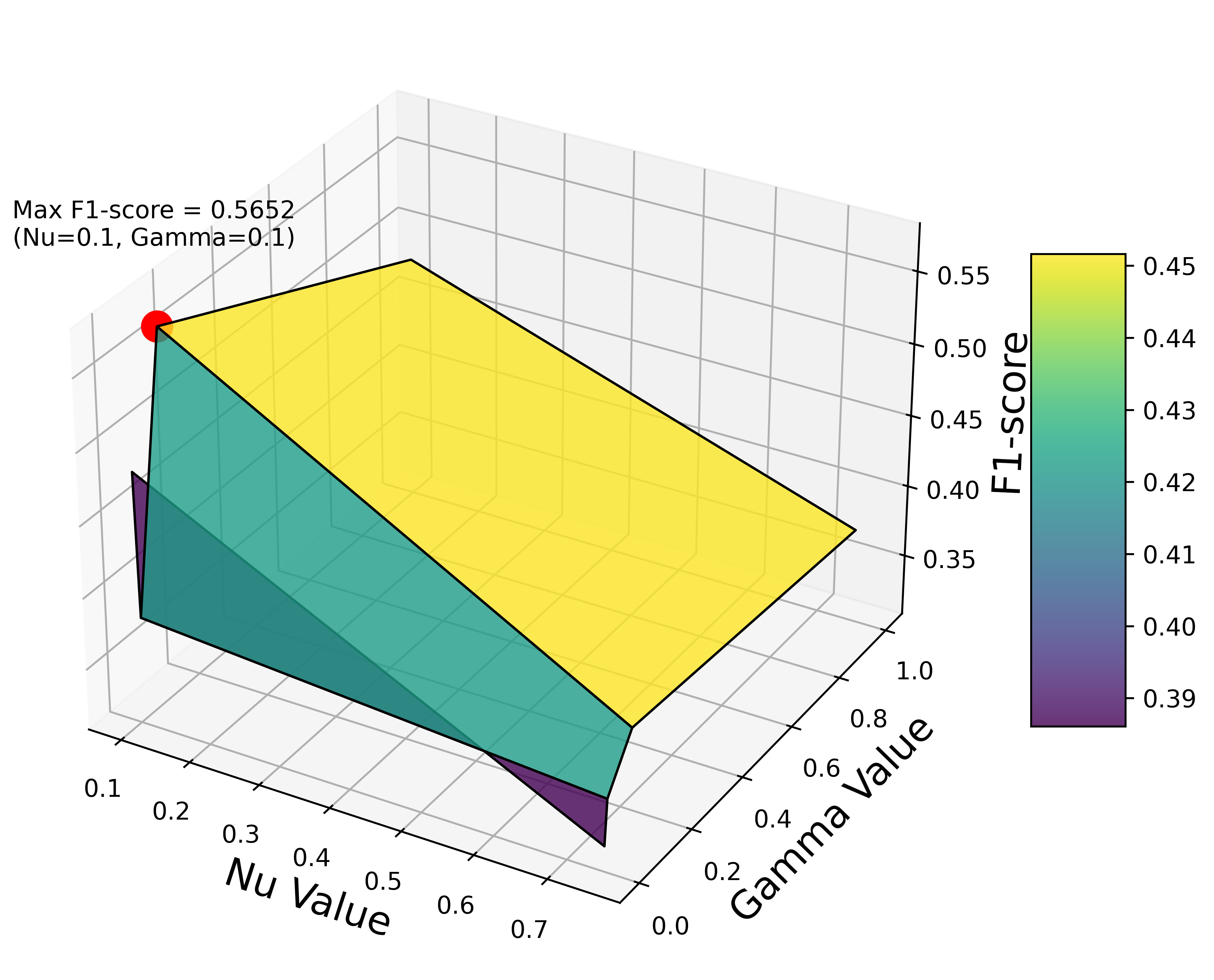}
        \caption{NuSVC}
    \end{subcaptionblock}
    \begin{subcaptionblock}{.45\textwidth}
        \centering
        \includegraphics[width=\textwidth]{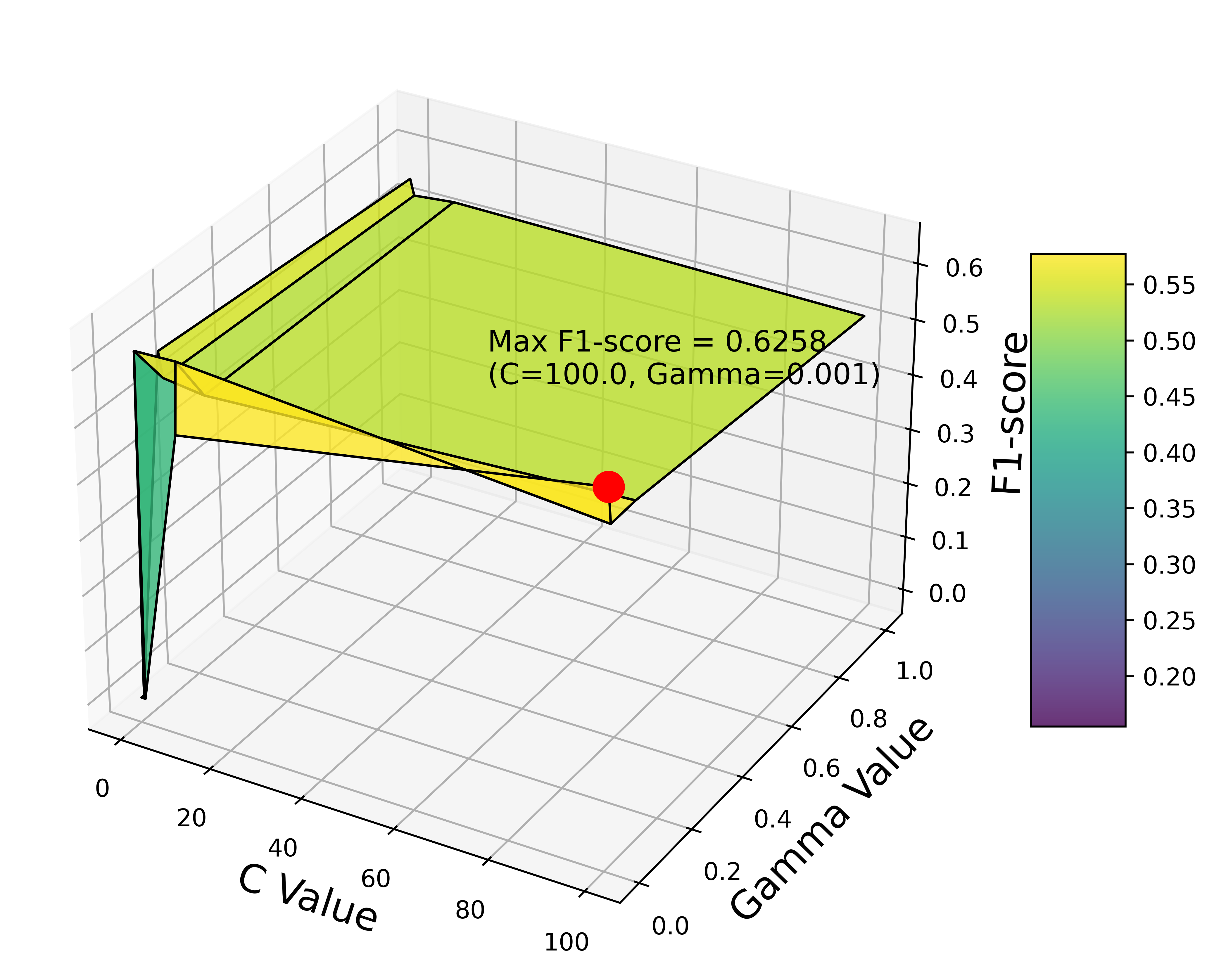}
        \caption{Proposed model}
    \end{subcaptionblock}

    \caption{Hyperparameter analysis for model performance using validation data and grid search in quake dataset.}
    \label{fig:fourplots}
\end{figure*}

\section{Discussion}
This study aimed to develop a novel SVM capable of handling class imbalance and noisy data scenarios. A new SVM formulation was introduced by incorporating a binary variable to account for misclassifications. The Benders decomposition technique was utilized to partition the original problem into a master problem and a subproblem, which were then solved iteratively.

\textbf{Hypothesis 1} was supported, demonstrating that adjusting the decision boundary in favor of the minority class effectively improves performance in imbalanced scenarios, addressing RQ1.
 
 In the proposed model, changing the representation of data provides flexibility in determining support vectors, similar to other research \cite{zhang2018robust}. When the minority class has a higher weight, the linear programming problem in the master problem prioritizes the inclusion of samples belonging to the minority class. Therefore, the decision boundary is shifted from the minority class toward the majority class.

In contrast, Soft Margin SVM applies the same misclassification cost to all training samples. In imbalanced datasets, majority class samples are often more densely distributed than minority class samples, even near the optimal decision boundary. Consequently, the ideal separating hyperplane may be affected by this imbalance \cite{batuwita2013class}, as observed in Table~\ref{tab:svm_comparison1} for the labor, fruitfly, and quake datasets.

Weighted SVM, which assigns different penalties to each class, demonstrated performance similar to that of Soft Margin  SVM (Table~\ref{tab:svm_comparison1}). This suggests that assigning equal weights to all samples within each class is insufficient to effectively address the class imbalance issue. Instead, samples, particularly those near the decision boundary, should be assigned varying significance, as implemented in the proposed model.

Similar to Weighted SVM, NuSVC allows for the incorporation of class weights to impose higher penalties on specific classes. However, its objective function lacks the flexibility to adjust the decision boundary in favor of the minority class. Consequently, it was not as effective as the proposed model.

The proposed model exhibited lower F1-score performance on the training set and higher performance on the test set (Table~\ref{tab:svm_comparison1}). The reduced performance on the training set arises from the model's iterative inclusion of samples near the decision boundary, with certain samples being excluded due to feasibility cuts imposed by Benders decomposition. This process mitigates overfitting to the training data, resulting in a more robust model that generalizes better on the test set.

It is important to note that the class imbalance ratio does not account for the distribution of data \cite{guan2022novel}. Data distribution is a critical factor for SVM, particularly near the decision boundary. Therefore, even in highly imbalanced scenarios, only the samples near the boundary between the two classes influence the decision boundary. The majority of samples that are distant from the boundary do not impact the determination of the decision function, despite contributing to the overall class imbalance ratio.

Hyperparameter analysis revealed that benchmark methods selected the lowest value (0.1) for \( C \) and \( \nu \), applying minimal penalties for margin violations in the quake dataset (Figure \ref{fig:fourplots}). This selection is attributed to the dataset's fraction of borderline points (0.47 in Table \ref{tab:dataset_stats}), indicating a high degree of class overlap near the decision boundary. In contrast, the proposed model selected the highest penalty (\( C \)=100), as it solves a hard margin SVM in the subproblem.

\textbf{Hypothesis 2} was supported, demonstrating that incrementally refining the decision boundary by updating support vectors enhances performance in noisy scenarios, addressing RQ2.

 The proposed model begins with a feasible solution in which both classes are perfectly separable. Samples are then added incrementally. At each iteration, the master problem selects a new sample from the candidate set by solving a linear programming problem. This newly introduced sample alters the data representation for the subproblem, which is subsequently solved with the updated dataset. The subproblem's solution generates new constraints that guide the selection of subsequent samples. This iterative process continues until convergence. Constructing the decision boundary through this mechanism resulted in superior performance compared to Soft Margin  SVM and NuSVC, which rely on penalizing samples based on the extent of violation.

The proposed model achieved the lowest number of support vectors across datasets compared to Soft Margin SVM and NuSVC (Figure \ref{fig:SVCount}). In the two datasets where NuSVC yielded a lower number of support vectors, it exhibited lower accuracy compared to the proposed model (cloud and rabe-266 datasets in Table~\ref{tab:svm_comparison}).
 The key advantage of the proposed model lies in its ability to dynamically modify the data representation and iteratively update support vectors (boundary shifting) by changing data representation, rather than relying on the initial data representation to select support vectors. This flexibility enables the proposed model to maintain the minimal number of required support vectors while shifting the boundary. The idea of boundary shifting has been utilized in neural networks using cross-entropy loss \cite{huang2023neural}; however, the novelty of this study is the presentation of a mathematical framework (master problem and subproblem) for updating the support vectors and shifting the boundary. 

The lower number of support vectors reduces the model's runtime memory usage, making the proposed model particularly suitable for resource-constrained environments (e.g., when using microcontrollers) \cite{saha2022machine}. In addition, the reduced prediction time (Table \ref{tab:train_pred_times}), resulting from the smaller set of support vectors, makes the proposed model highly effective for real-world applications such as prosthesis control, activity recognition, and fall detection \cite{zhang2021class}, where both real-time prediction constraints and class imbalance pose challenges for ML models.

While the reduced prediction time is a significant advantage, it comes at the cost of increased training time. The iterative approach to solving the problem results in prolonged training time (Table \ref{tab:train_pred_times}), as the model trains multiple SVM models until convergence. Nevertheless, the training time remains manageable for practical applications. Another limitation of the proposed model is that its effectiveness relies on scenarios where class boundaries exhibit overlap. Despite this, the model offers several advantages, including the provision of a unique optimal solution, a reduced number of support vectors, and robustness to noise and outliers.

Future work includes expanding the proposed model for support vector regression, conducting sensitivity analysis on noise levels and boundary complexity, evaluating the impact of different kernel functions, and exploring its applicability to multiclass or larger datasets.

\section{Conclusion}

This study demonstrated that an SVM model that penalizes the number of misclassified samples is more effective than the conventional Soft Margin SVM in handling class imbalance and noisy data. The proposed model incorporates a binary variable in the objective function, formulating the problem as a mixed-integer programming task. It is then solved using the Benders decomposition technique. By decomposing the problem into a master problem and a subproblem, the optimal solution is obtained iteratively. The model’s effectiveness was evaluated across multiple datasets with class imbalance and noise. Experimental results indicated that the proposed model outperforms both the Soft Margin SVM and NuSVC. It effectively addressed class imbalance by adjusting the decision boundary in favor of the minority class. In addition, it demonstrated robustness to noisy data by disregarding outliers. The proposed model utilized fewer support vectors for boundary determination, reducing prediction time and enhancing its practicality for real-world applications. To facilitate adoption, an open-source Python implementation of the proposed model is available for use in various classification tasks\footnote{https://github.com/MojtabaMohasel/BSVM.git}.

\section{Acknowledgment}
Computational efforts were performed on the Tempest High Performance Computing System, operated and supported by University Information Technology Research Cyberinfrastructure at Montana State University.

\bibliographystyle{IEEEtran}

\bibliography{paper}

\end{document}